\documentclass{article} % For LaTeX2e
\usepackage{iclr2024_conference,times}

\usepackage{amsmath,amsfonts,bm}

\def\eqref#1{equation~\ref{#1}}
\def\1{\bm{1}}

\DeclareMathAlphabet{\mathsfit}{\encodingdefault}{\sfdefault}{m}{sl}
\SetMathAlphabet{\mathsfit}{bold}{\encodingdefault}{\sfdefault}{bx}{n}

\usepackage{booktabs}           % professional-quality tables
\usepackage{multirow}           % tabular cells spanning multiple rows
\usepackage{amsfonts}           % blackboard math symbols
\usepackage{graphicx}           % figures
\usepackage{duckuments}
\usepackage{mathtools}          % coloneqq
\PassOptionsToPackage{hyphens}{url}
\usepackage{hyperref}

\usepackage[linesnumbered,vlined,ruled,commentsnumbered]{algorithm2e}
\usepackage[inline]{enumitem}

\title{Learning Fair Representations with High-Confidence Guarantees}
\newcommand{\proj}{FRG\xspace}
\newtheorem{theorem}{Theorem}[section]

\newtheorem{definition}[theorem]{Definition}

\newtheorem{lemma}[theorem]{Lemma}
\newtheorem{property}[theorem]{Property}

\author{Yuhong Luo \\
University of Massachusetts, USA\\
\texttt{yuhongluo@umass.edu} \\
\And
Austin Hoag \\
Berkeley Existential Risk Initiative, USA\\
\texttt{austinthomashoag@gmail.com} \\
\And
Philip S. Thomas \\
University of Massachusetts, USA\\
\texttt{pthomas@cs.umass.edu} \\
}

\iclrfinalcopy % Uncomment for camera-ready version, but NOT for submission.
\begin{document}

\maketitle

\begin{abstract}
Representation learning is increasingly employed to generate representations that are predictive across multiple downstream tasks.
The development of representation learning algorithms that provide strong fairness guarantees is thus important because it can prevent unfairness towards disadvantaged groups for \emph{all} downstream prediction tasks. %To prevent unfairness towards disadvantaged groups in all downstream tasks, it is crucial to provide representation learning algorithms that provide fairness guarantees.
In this paper, we formally define the problem of learning representations that are fair with high confidence.
We then introduce the \emph{\textbf{F}air \textbf{R}epresentation learning with high-confidence \textbf{G}uarantees} (\proj) framework, which provides high-confidence guarantees for limiting unfairness across all downstream models and tasks, with user-defined upper bounds.
After proving that \proj ensures fairness for all downstream models and tasks with high probability, we present empirical evaluations that demonstrate \proj's effectiveness at upper bounding unfairness for multiple downstream models and tasks.
\end{abstract}

% \vspace{-0.5cm}
\section{Introduction}
\vspace{-0.2cm}

%The primary goal of \emph{machine learning} (ML) algorithms is to make accurate predictions. 
%
In every prediction task, machine learning (ML) algorithms assume two distinct roles: the data producer and the data consumer~\citep{Zemel2023Learning, Dwork2012Fairness, Madras2018Learning}.
The data consumer's role is to make accurate predictions using the data provided by the data producer.
While the data producer may distribute raw data, it is also common for it to generate new representations for the input data (referred to as \emph{representation learning}) that are predictive.
% %
% In the context of representation learning, the data producer's role is to generate data representations that are , while the data consumer's role is to make accurate predictions using these representations.
% consider two distinct roles within every prediction task: 

% This task can sometimes be split into two parts and each : 
%
When multiple data consumers' prediction tasks involve inputs of the same type, such as natural language text or images, the data producer can generate \emph{general representations} that are predictive to multiple subsequent tasks.
With the increasing prominence of deep learning methods for tasks involving language, audio, and visual data, the adoption of these general representations has become prevalent in both academic and industrial settings.
Examples of this trend include the Variational Autoencoder (VAE)~\citep{kingma2013autoencoding} and recent language models such as BERT~\citep{bert2019devlin} and GPT-4~\citep{GPT2023openai}.
While representation learning can benefit various downstream predictions, it is also susceptible to the risk of producing unintended or undesirable behaviors in the downstream tasks. %Representation learning is a powerful tool for transforming input data into condensed representations that can benefit various downstream predictions. However, these representations are susceptible to the risk of producing unintended and undesirable behaviors in the downstream tasks.
More precisely,  the representations can be used to generate predictions that display unfairness or bias against certain disadvantaged groups.
%
%\textcolor{red}{AH: the following sentence is too long. Suggested rewording: 
This is especially problematic in critical domains, such as loan underwriting~\citep{Byrnes16}, hiring~\citep{Miller15} and criminal sentencing~\citep{Angwin16}, where the consequences of algorithmic bias may severely impact individuals. In these cases, fairness may mean that a person's sensitive attributes, such as race and gender, should
not be pertinent to the model's predictions.  %} For instance, when a representation is applied to downstream predictions in critical domains like financial loan approval~\citep{Byrnes16}, hiring~\citep{Miller15} and criminal sentencing~\citep{Angwin16}, and if the subsequent models exhibit algorithmic bias, individuals may be profoundly affected due to certain \emph{sensitive attributes} such as race and gender, which ideally should not be pertinent to the predictions.
Given these risks, researchers propose that fairness should be a concern not only for the data consumer that uses the representations, but also for the data producer that generates them \citep{Zemel2023Learning, Madras2018Learning}.
%
%\textcolor{red}{AH: As we will show,} 
A data producer that ensures fairness in the data representation ensures fairness in all downstream tasks. As a result, the data producer can release the data representation to any data consumer without concern.

To ensure the fairness of representations, simply removing sensitive attributes from the dataset is insufficient as sensitive information can still inadvertently leak through non-sensitive attributes~\cite{Castelnovo2022a}.
Numerous studies have proposed methods for learning fair representations, collectively referred to as \emph{fair representation learning}~\citep{Zemel2023Learning}.
These methods are designed to reduce the presence of sensitive information in the representations, ensuring fairness across all downstream models and tasks. 
In Section \ref{sec:rw} we present a detailed discussion of related work such as the works of \citet{Louizos2016Variational, Madras2018Learning,Moyer2018Invariant,Song2019controllable,Gupta2021Controllable, Balunovic2022Fair,Kim2022Learning} and \citet{Jovanovic2023FARE}.
%
% Many of these methods also provide theoretical upper bounds on the unfairness of all downstream models, quantified by metrics such as demographic parity, equalized odds, and equal opportunity~\citep{Dwork2012Fairness,Hardt2016Equality}.
% %
% In addition, they propose algorithms aimed at minimizing these upper bounds, effectively reducing unfairness(e.g.,~\citep{Gupta2021Controllable, Madras2018Learning}).

While much of the previous work has demonstrated effectiveness in promoting fairness in specific downstream tasks, the majority of these approaches provide little to no assurance that the unfairness of all downstream models will be consistently controlled or bounded by a user-defined error threshold with high probability.
In many areas of supervised learning, providing high-confidence guarantees is considered essential for ensuring fairness, privacy, and safety of the learning algorithm 
~\citep{Li2022Fairee, Abadi2016Deep, Philip2019Preventing}.
%While many areas of study within supervised learning consider the provision of a high-confidence guarantee  crucial for ensuring the fairness, privacy, and safety of the learning algorithm
This need for high-confidence guarantees becomes even more critical in the context of learning fair representation as the absence of such guarantees can lead to undesired behaviors across multiple downstream applications.
% in fair representation learning can result in 

% To our knowledge, only one line of research in fair representation learning provides practical certificates that upper bound the unfairness, specifically measured by demographic parity, with high probability for all downstream models~\citep{Balunovic2022Fair,Jovanovic2023FARE}.
% %
% However, they do not provide a framework for learning and producing fair representation models with high confidence. Specifically, they do not accept an error threshold specified by the user and ensure with high confidence that the unfairness of all downstream models is upper-bounded by that threshold.
% %
% Their methods are also limited to particular representation model architectures (e.g., decision trees, normalizing flows) and representation distributions (e.g., discrete), which may not be ideal in some situations.

%In this paper, we introduce a framework that offers a high-confidence fairness guarantee for representation learning.
%, compatible with most parameterized representation models.
%
 Our main contributions in this paper are as follows.
 We provide formal definition of learning representations that are fair with high confidence in Section~\ref{sec:problem}. This definition ensures that unfairness, measured using demographic parity,
 is consistently upper-bounded by a user-defined threshold across all downstream predictions with high probability.
Under this definition, we introduce the \emph{\textbf{F}air \textbf{R}epresentation learning with high-confidence \textbf{G}uarantees} (\proj) framework in Section~\ref{sec:method}.
 % that provides high-confidence fairness guarantees
 %
 After proving that \proj ensures fairness for all downstream models and tasks with high probability in Section~\ref{sec:theory},
 %We provide theoretical guarantees  that \proj ensures fairness according to demographic parity in .
 %
we present empirical evaluations that demonstrate \proj's effectiveness at upper bounding unfairness for multiple downstream models and tasks in Section~\ref{sec:exp}.
\vspace{-0.12cm}
\section{Background}\label{sec:background}
\vspace{-0.12cm}

% notation table
% \begin{figure*}
%     \centering
    % \vspace{-2.5mm}

%\begin{table}[t]
%\scriptsize \centering
% \resizebox{1\textwidth}{!}{
% \begin{tabular}{cllll}
% \hline
% \textbf{No.} & \textbf{Notations} & \multicolumn{1}{c}{Definitions} \\ \hline
% 1. &   & A representation learning algorithm $a$ \\\input{03-methodology}

% 1. & $g$        & The constraint function measures undesirable behavior. When $g(\theta) > 0$, the constraint is violated.\\  %and could outputs undesirable behavior

% 3. &   & \\
% 4. &   &  \\
% 5. &    & .\\
% \hline \end{tabular}}
% \end{figure*}
% 1. Introduce representation learning from the perspective of information theory.\\
% Adversarial objective.\\
%\TODO{Define the notion of fairness of a model. A model is only fair or unfair. If it satisfy the constraint then fair. The algorithm that produces a fair model with high confidence is a Seldonian algorithm. }
%This section first introduces the notations of a representation learning method and the fairness requirement for a representation learning model to be unbiased. Then we formally define our goal in p\simroviding high confidence guarantee for an algorithm, such that any output model satisfies the desired fairness requirement with high probability.\\
In this section,
we introduce the notation used for representation learning, define a measure of unfairness for classification models, and review a useful property relevant to fair representation learning.
%used in a representation learning method and the fairness requirement that a representation learning model must meet to be unbiased. We then proceed to formally define our objective of providing a high confidence guarantee for a representation learning algorithm. This guarantee ensures that any output model satisfies the desired fairness requirement with a high probability.
\vspace{-0.4cm}
\subsection{Notation for Representation Learning}
\vspace{-0.15cm}
Let $X$ be a random variable called the \emph{feature vector}, $S$ be a random variable called the  \emph{sensitive attribute}, and $D\coloneqq\{(X_i, S_i)\}^n_{i=1}$ be the \emph{dataset}, where $X_1 \ldots X_n$ are i.i.d.~random variables with the same distribution as $X$, $S_1 \ldots S_n$ are i.i.d.~random variables with the same distribution as $S$, and each $(X_i, S_i)$ has the same joint distribution as $(X,S)$.
%, and $\forall_i (X_i, S_i)$ has the same joint distribution as $(X,S)$.
We define $\mathcal{D}$ to be the set of all possible $D$.
Let $\phi \in \Phi$ be the  \emph{representation model parameters} and $q_{\phi}$ be the  \emph{representation model} parameterized by $\phi$.
Then, we define $Z$ to be the  \emph{representation} for $(X,S)$ where $Z\sim q_{\phi}(\cdot|X,S)$ and $Z\in {\mathbb{R}^l}$.

We assume that the learned representation will be used for subsequent supervised learning tasks, which we call \emph{downstream tasks}. We denote the \emph{label} for such a downstream task as the random variable $Y$. %We define a downstream task as follows. Let $Y$ be the \emph{downstream task label}.
The objective in a downstream task is to predict $Y$ given $(X,S)$. Instead of using $(X,S)$ directly as input, we use $Z$ as input to a \emph{downstream model} $\tau: \mathbb{R}^l \rightarrow \mathbb{R}$.
Let $\hat{Y} \coloneqq \tau(Z)$ denote the prediction of $Y$ produced by model $\tau$. We call $\hat Y$ the \emph{downstream prediction}. 
Notice that different downstream tasks correspond to different joint distributions of $(X,S,Y)$, but we assume all downstream tasks share the same joint distribution of $(X,S)$. Thus, the same representation $Z$ can be used for multiple downstream tasks. 

% $\mathcal{D}$ be the set of all possible datasets that can be used for training a representation learning model and $D \in \mathcal{D}$ be any given dataset. Then $D =\{(x_i, s_i)\}^N_{i=1}$  contains arbitrary feature vectors $x\in \mathcal{X}$  and sensitive or protected attributes  $s \in \mathcal{S}$. We assume each of the data points is sampled i.i.d. from an unknown distribution. Let  $z \in {\mathbb{R}^l}$ be a representation of dimension $l$ for a data point $(x, s) \in D$ with a prior distribution $p(z)$ which is generally a multivariate Gaussian distribution. We assume $z$ for a single data point $(x, s)$ can be obtain by sampling from a posterior distribution $q_{\phi}(z|x, s)$ where $\phi \in \Phi$ is a parameterized model. Let $\tau \in \mathcal{T}: \mathbb{R}^l 
% \rightarrow \mathbb{R}$ be any downstream decision model that only relies on representation $z$ to make a prediction. Given a representation $z$, $\tau(z) = \hat{y}$ is the predicted label.
\vspace{-0.2cm}
\subsection{A Measure of Unfairness for Downstream Models}
\vspace{-0.15cm}
We will define a fair representation model to be one that ensures fairness for all possible downstream tasks and all possible downstream models. To achieve this, we must first establish a definition of fairness for downstream models and tasks.
%Before we define fairness for a representation model, we start by defining the measure of unfairness for  downstream models.
%
In this work, we focus on classification tasks and a widely used group fairness measure based on demographic parity~\citep{Dwork2012Fairness}.
Let $Y$ and $S$ be binary, leading to the following definition.
%
% PST: Say that different downstream tasks correspond to different joint distributions of (X,S,Y), but note that we assume all downstream tasks share the same joint distribution of (X,S). Then in Definition 3.2 you can say "for any downstream task and any downstream model $\tau$ for that task."
%
\begin{definition}
    [A measure of how unfair a downstream model $\tau$ is under demographic parity] Let $\Delta_\text{DP}(\tau,\phi)$ denote a measure of how unfair downstream predictions $\hat Y$ are when using representation parameters $\phi$ and downstream model $\tau$. Specifically, 
    \begin{align}\label{eq:dp}
        \Delta_\text{DP}(\tau, \phi) \coloneqq \Big |\Pr(\hat{Y} = 1 | S = 1) -  \Pr(\hat{Y} = 1 | S = 0)\Big |.
    \end{align} 
\end{definition}
When $S$ is non-binary, $\Delta_\text{DP}(\tau, \phi)$ is defined as the maximum absolute difference between the conditional probabilities, $\Pr(\hat{Y} = 1 | S)$, with any pair of values of $S$~\citep{bird2020fairlearn}.
% \textcolor{red}{Phil: I'm not sure, but I think in the previous sentence ``implemented'' isn't really what you mean. Perhaps cut everything in the parenthesis and just end with \citep{bird2020fairlearn}?}

\vspace{-0.2cm}
\subsection{Mutual Information Bounds Demographic Parity}
\vspace{-0.15cm}
The demographic-parity-based measure  is specified for downstream models.
However, we want our representation model to guarantee fairness for every possible downstream model and downstream task.
\citet{Gupta2021Controllable} derived a bound for $\Delta_\text{DP}(\tau, \phi)$ that removes the dependency on the downstream model $\tau$.
Specifically, \citet{Gupta2021Controllable} showed  %Property~\ref{eq:dpbound} 
that the mutual information between the representation and the sensitive attributes, denoted by $I(Z;S)$, can be used to limit the demographic parity of downstream models. 
\begin{property}[Relation of mutual information to $\Delta_\text{DP}(\tau, \phi)$]\label{eq:dpbound}
For all downstream models $\tau$ in all downstream tasks,  $$I(Z;S) \ge \psi(\Delta_\text{DP}(\tau, \phi)),$$
where $\psi$ is a strictly increasing non-negative convex function derived by \cite{Gupta2021Controllable}, and the details of which are in Appendix~\ref{apx:psi}. 
\textbf{Proof.} See the work of \cite{Gupta2021Controllable}.
\end{property}

Notice that Property~\ref{eq:dpbound} does not provide a direct upper bound on $\Delta_\text{DP}(\tau,\phi)$. Instead, it provides an upper bound on a strictly increasing non-negative convex function of $\Delta_\text{DP}(\tau,\phi)$. We use this property later to guarantee fairness for representation models with high confidence (see Sec.~\ref{sec:method}).

\vspace{-0.2cm}
\section{Problem Statement} %: Representation Learning with High-Confidence Fairness Guarantees}
~\label{sec:problem}
\vspace{-0.5cm}
%In this section, we formally state our key objective of providing a representation learning algorithm with high confidence fairness guarantee.
%

In this section, we define what it means for a representation model to be fair. We then formulate the problem of learning representation models with high-confidence fairness guarantees.
\vspace{-0.1cm}
\subsection{The Definition of Fair Representation Models}
\vspace{-0.1cm}
A fair representation model should ensure with high confidence that the representations it generates will not lead to unfairness for downstream tasks. Specifically, a representation model is fair if and only if it results in fair predictions (as defined in Def.~\ref{eq:dp}) for every possible downstream model and downstream task. That is, for all downstream tasks and all $\tau$, $\Delta_{\text{DP}}(\tau, \phi)$ must be upper-bounded by a small constant, $\epsilon$.
Formally, we define an  ``$\epsilon$-fair" representation model as follows.
\begin{definition}
    [``$\epsilon$-fair" representation model]\label{eq:fair_model}
    Representation model $q_{\phi}$ is ``$\epsilon$-fair" with parameter $\epsilon \in [0,1]$ if and only if $\Delta_{\text{DP}}(\tau, \phi) \le \epsilon$, for every downstream model $\tau$ and downstream task.
\end{definition}
\vspace{-0.15cm}
\subsection{Problem Formulation}
\vspace{-0.1cm}
We define a representation learning algorithm $a: \mathcal{D} \rightarrow \Phi$ to be an algorithm that takes as input a data set and produces as output representation model parameters.
In this paper, we aim to provide a representation learning algorithm such that any representation model it learns is guaranteed to be $\epsilon$-fair under Def.~\ref{eq:fair_model}, with high confidence.
Such an algorithm has the following formal definition.
\begin{definition}[A representation learning algorithm with high-confidence fairness guarantees]
\label{def:highConfFairRep}
Given $\epsilon \in [0,1],\delta \in (0,1)$, and a dataset $D$, a representation learning algorithm $a$ is said to provide a $1-\delta$ confidence ``$\epsilon$-fairness" guarantee if and only if 
\begin{align}
   \Pr\left(g_{\epsilon}(a(D)) \le 0 \right) \ge 1 - \delta, 
\end{align}
where $g_{\epsilon}(\phi) \coloneqq  \sup_\tau \Delta_{\text{DP}}(\tau, \phi) - \epsilon$. %and $\delta \in (0,1)$ such that $1-\delta$ is the target confidence level.
%
%\textcolor{blue}{Note: Phil thinks we should have $g_\epsilon(\phi)=\max_\tau \Delta_\text{DP}(\tau,\phi) - \epsilon.$ My hesitation is that here ``max'' isn't great. Perhaps this should be ``sup'', but that will probably confuse more readers than it will help. Thinking more, I now think we should use $\sup_\tau$ instead of $\max_\tau$.}
%
\label{eq:seldonian}
\end{definition}
%Observe that if $g_{\epsilon}(\phi) > 0$, the constraint on mutual information is violated and the representation learning model is biased. Then, our goal is to create an algorithm $a$ with the following probabilistic guarantee: 

Observe that $q_{\phi}$ is an $\epsilon$-fair representation model if and only if $g_{\epsilon}(\phi) \le 0$ (Def.~\ref{eq:fair_model}).
Therefore, any algorithm under Def.~\ref{eq:seldonian} guarantees that any representation model with parameters learned by this algorithm has at least $1-\delta$ probability to be an $\epsilon$-fair representation model.

According to \cite{Philip2019Preventing}, such algorithms can be categorized as \textit{Seldonian} algorithms.

%where $\delta \in [0,1]$ is the require$\Delta_\text{DP}(\tau)$d confidence level. Notice that $D$ is the only random variable for this problem. 

\textbf{Special case: unachievable $\epsilon$-fair representation models.}
In some scenarios it may not be possible for any non-degenerate algorithm to ensure fairness with the specified confidence $1-\delta$, for example when $\epsilon$, $\delta$, and the amount of available training data are all very small. In such cases, we allow the algorithm to output \emph{No Solution Found} (\texttt{NSF}) as a way of indicating that it is unable to provide the required confidence that the learned representation will be fair given the amount of data it has been provided. To indicate that it is always fair for the algorithm to return \texttt{NSF}, we define $g_\epsilon(\phi)=0$. 
% In some scenarios, for instance, when the data size is not large enough or when $\epsilon$ is very small, there may not exist, or the algorithm may not find any $\phi$ such that $g_{\epsilon}(\phi) \le 0$.
%
% Therefore, we allow a Seldonian algorithm to output \emph{No Solution Found} (\texttt{NSF}), and we define $g_{\epsilon}(\texttt{NSF}) = 0$.
% %
%Thus, NSF is considered a fair output of an algorithm.
%
However, if an algorithm constantly returns \texttt{NSF}, it is of no value.
We empirically evaluate the probability of returning a solution (i.e., not \texttt{NSF}) for our algorithm in  Section \ref{sec:exp}.
\begin{figure*}
\centering
\vspace{-5.0mm}
\includegraphics[trim={0.3cm 0.3cm 0.1cm 0.3cm}, clip,width=0.65\textwidth]{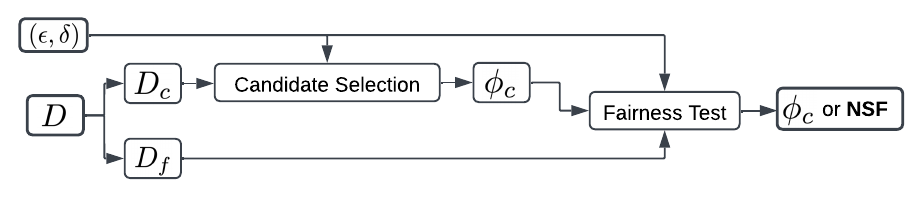}
\vspace{-3.7mm}
%\hspace{-3mm}
\caption{\small{
An overview of \proj. See Sec.~\ref{sec:method} for discussion.
}}
\label{fig:alg}
\vspace{-2.0mm}
\end{figure*}

\vspace{-2.0mm}
\section{Methodology}\label{sec:method}
In this section, we introduce the framework \emph{\textbf{F}air \textbf{R}epresentation learning with high-confidence \textbf{G}uarantees} (\proj). It is the first representation learning algorithm that provides the high-confidence fairness guarantee specified in Def.~\ref{def:highConfFairRep}. An overview of \proj is provided in Fig.~\ref{fig:alg}.
% \proj, the novel representation learning framework that guarantees fairness with high confidence. Specifically, we propose a Seldonian algorithm that follows Eq.~\ref{eq:seldonian} and outputs models that satisfy the constrained objective in Eq.~\ref{eq:objective} with high probability.

\proj consists of two major components called \emph{candidate selection} and the \emph{fairness test}.
First we present a high-level summary of the algorithm before discussing each component in more detail. %The high-level view of the algorithm is as follows.
\proj first splits the data $D$ into disjoint sets, $D_c$ and $D_f$.
%, which are used by the candidate selection and the fairness test respectively.
%
Candidate selection uses $D_c$ to optimize and propose 
\emph{candidate solution} $\phi_c$, while the fairness test uses $D_f$ to evaluate whether $\phi_c$ can satisfy $g_{\epsilon}(\phi_c) \le 0$ (Def.~\ref{eq:seldonian})  on future unseen data with sufficient confidence.
Finally, \proj returns $\phi_c$ or \texttt{NSF} according to the result of the fairness test.
Notice that a candidate selection algorithm that does not consider fairness may often propose candidate solutions that fail the fairness test, resulting in \texttt{NSF}.
In the following subsections, we  introduce the fairness test and then provide details for a candidate selection algorithm that %.
%
%Then, we describe our design of candidate selection which 
%learns to 
proposes candidate solutions that  generate representations of high expressiveness and that are likely to pass the fairness test. %s with high likelihood.
%
%We summarize the algorithm in Alg.
\vspace{-0.2cm}
\subsection{Fairness Test}\label{sec:ft}
% \vspace{-0.1cm}
The goal of the fairness test is to evaluate whether a candidate solution $\phi_c$ induces a fair representation model with high confidence.
%
%\textcolor{red}{This sounds like we aren't doing $\epsilon$-fairness, when we still are. Instead say this is how we ensure epsilon fairness, not an alternative.}
In this section, we first develop $\Tilde{g}_\epsilon(\phi)$ where $\Tilde{g}_{\epsilon}(\phi) \le g_{\epsilon}(\phi)$, and propose evaluating $\Tilde{g}_{\epsilon}(\phi) \le 0$ to provably determine whether a representation model $\phi$ is $\epsilon$-fair, i.e., $g_{\epsilon}(\phi) \le 0$. %\textcolor{red}{[Phil: What does this mean? It's tractable... but just an estimate? Does theory not apply? Is it the case that $\tilde g(\phi) \leq g(\phi)$ always? Explain this to the reader -- try not to allow for any confusion or uncertainty in their mind as they read.]} % with high a tractable alternative measure for $\epsilon$-fairnes to the demorgraphic-parity-based $g_{\epsilon}(\phi)$.
We then propose the construction of a high-confidence upper bound on $\Tilde{g}_{\epsilon}(\phi)$.
We finally detail the evaluation process for a candidate solution $\phi_c$ using this high-confidence upper bound.

%\phil{We compute a $1-\delta$ confidence prediction of whether the model is $\epsilon$-fair. We do this by computing a $1-\delta$ confidence upper bound on $\Delta_\text{DP}(\tau,\phi)$. I don't think it's right to talk about an ``upper bound'' on $\epsilon-$fairness, since epsilon fairness is a Boolean property that models either have or do not have.}

\textbf{A mutual-information-based evaluation.}
Our goal is to evaluate whether $g_{\epsilon}(\phi) \le 0$ with high confidence. However, estimating $g_{\epsilon}(\phi)=\sup_\tau \Delta_{\text{DP}}(\tau, \phi) - \epsilon$ is intractable because it requires knowledge of all downstream models and all downstream tasks.
To remove the dependency on downstream models, we apply Property~\ref{eq:dpbound}, and evaluate whether $ I(Z;S) - \psi(\epsilon) \le 0$ instead of $\sup_\tau \Delta_{\text{DP}}(\tau, \phi) - \epsilon \le 0$ ($\psi$ is derived by \citet{Gupta2021Controllable} and defined in Appendix \ref{apx:psi}).
%, to replace the evaluation $g_{\epsilon}(\phi) \le 0$.
%
Intuitively, when the mutual information between the representation and the sensitive attribute is low, it is hard for any model to predict $S$ given $Z$ with high accuracy.
Therefore, any downstream model that does not explicitly aim to predict $S$ is even less likely to take advantage of the sensitive attribute to produce unfair predictions.
Theoretically, evaluating  $ I(Z;S) - \psi(\epsilon) \le 0$ can provably determine the $\epsilon$-fairness of $\phi$ under Def.~\ref{eq:fair_model}.
%\yuhong{Original sentence was ``Theoretically, these two evaluations are equivalent for determining $\epsilon$-fairness of $\phi$." (This sentence should be wrong. The proxy evaluation implies the original evaluation. Changing to the sentence before this.)}
%
We postpone the theoretical analysis to Sec.~\ref{sec:theory}.

Unfortunately, %it turns out that
computing $I(Z;S)$ is intractable because it requires marginalizing the joint distribution of $(X,S,Z)$ over feature vector $X$, and so even this approach remains intractable.
Multiple previous works have derived tractable upper bounds on $I(Z;S)$, which we discuss in detail in Appendix~\ref{apx:upperbounds}. % including the works of \citet{Song2019controllable, Moyer2018Invariant, Gupta2021Controllable}, and we discuss these upper bounds in detail in Section~\ref{sec:upperbounds}.
Let $\Tilde{I}(Z;S)$ be one of these tractable upper bounds on $I(Z;S)$.
Then, we define 
\begin{align}\label{eq:g_tilde}
  \Tilde{g}_{\epsilon}(\phi) \coloneqq \Tilde{I}(Z;S) - \psi(\epsilon).  
\end{align}
With this upper bound, we now evaluate the $\epsilon$-fairness of $\phi$ by evaluating $\Tilde{g}_{\epsilon}(\phi) \le 0$. In Lemma~\ref{theorem:downstream}, we prove  if $\Pr\left(\Tilde{g}_{\epsilon}(a(D)) \le 0 \right) \ge 1 - \delta$, then algorithm $a$ provides the desired high-confidence fairness guarantee.% in Def.~\ref{eq:fair_model}.

\textbf{$1-\delta$ confidence upper bound on $\Tilde{g}_{\epsilon}(\phi)$.}
We follow two steps to compute a $1-\delta$ confidence upper bound on $\Tilde{g}_{\epsilon}(\phi)$ (if this high-confidence upper bound is at most zero, then we can conclude that $\Tilde{g}_\epsilon(\phi) \leq 0$ with confidence $1-\delta$).
%We evaluate the probability of $\Tilde{g}_{\epsilon}(\phi) \le 0$ through a construction of the  $1-\delta$ confidence  upper bound on $\Tilde{g}_{\epsilon}(\phi)$. We follow these two steps.
%
(1) Obtain $m$ i.i.d.~unbiased estimates $\hat{g}^{(1)}, \ldots, \hat{g}^{(m)}$ of $\Tilde{g}_{\epsilon}(\phi)$ using $D_f$, i.e., $\mathbb{E}[\hat{g}^{(j)}] = \Tilde{g}_{\epsilon}(\phi)$ for any $j\in[1,...,m]$.
(2) Apply standard statistical tools such as Student's t-test~\citep{studentttest} or Hoeffding's inequality~\citep{Hoeffding}  to construct a $1-\delta$ confidence upper bound on $\Tilde{g}_{\epsilon}(\phi)$ using $\hat{g}^{(1)}, \ldots, \hat{g}^{(m)}$. Note that we use Student's t-test for our experiments (Sec.~\ref{sec:exp}).

We define $U_{\epsilon}: (\Phi, \mathcal{D}) \rightarrow \mathbb{R}$ to be such a function that produces a $1-\delta$ confidence upper bound. Specifically, for $U_{\epsilon} (\phi,D_f)$, we have the following,
% constructs a $1-\delta$ confidence upper bound for $g_{\epsilon}(\phi)$.
%
\begin{align}
    \text{Pr}\Big(\Tilde{g}_{\epsilon}(\phi) \le U_{\epsilon}(\phi, D_f) \Big) \ge 1 - \delta.
\label{eq:confidence}
\end{align}
\textbf{Evaluation of candidate solutions.} Suppose the fairness test gets a candidate solution $\phi_c$ and  $U_{\epsilon} (\phi_c,D_f) \le 0$, it follows that there is at least confidence $1-\delta$ that $\Tilde{g}_{\epsilon}(\phi_c) \le 0$ (Inequality~\ref{eq:confidence}).
Then, the fairness test concludes with at least $1-
\delta$ confidence that $q_{\phi_c}$ is an $\epsilon$-fair representation model, and $\phi_c$ passes the test.
If, however, $U_{\epsilon} (\phi_c,D_f) > 0$, then the algorithm cannot conclude that $g_{\epsilon}(\phi_c) \le 0$ with high confidence. 
Therefore, the fairness test concludes that there is not sufficient confidence that $q_{\phi_c}$ is an $\epsilon$-fair representation model, and $\phi_c$ fails the test.

Finally, if $\phi_c$ passes the fairness test, \proj outputs $\phi_c$. Otherwise, it outputs \texttt{NSF}. When $\phi_c$ fails the fairness test, we do not search for and test another representation model because this would result in the well known ``multiple comparisons problem.'' In this case, each run of the fairness test can be viewed as a hypothesis test for determining whether the representation is fair with sufficient confidence. 

\vspace{-0.12cm}
\subsection{Candidate Selection}\label{sec:cs}
Notice that a representation learning algorithm using the fairness test mechanism as designed in Sec.~\ref{sec:ft} provides the desired $1-\delta$ confidence $\epsilon$-fairness guarantee (Def.~\ref{eq:seldonian}) regardless of the choice of candidate selection, as shown in Thm.~\ref{theorem:conf}.
However, candidate selection is considered ineffective if most of its proposed solutions fail the fairness test.
In this section, we introduce an effective design for candidate selection. This design results in candidate solutions that are both likely to pass the fairness test and optimized for high expressiveness.

\subsubsection{Predicting Whether a Candidate Solution Will Pass the Fairness Test}
The candidate selection mechanism proposes a candidate solution $\phi_c$ that it predicts will pass the fairness test.
Such a prediction can be naturally made by leveraging knowledge of the exact form of the fairness test, except using dataset $D_c$ instead of $D_f$, i.e., checking whether $U_{\epsilon}(\phi_c, D_c) \le 0$. However, there is one caveat.
We repeatedly use the same dataset $D_c$ to construct high confidence upper bounds while searching for the candidate solution.
Therefore, we may overfit to $D_c$, resulting in an overestimation of the confidence that the candidate solution will pass the fairness test.
To mitigate this issue, we inflate (double the width of) the confidence interval used in candidate selection. % for the construction of the upper bound.
We denote the inflated $1-\delta$ confidence upper bound on $\Tilde{g}_{\epsilon}(\phi_c)$ as $\hat{U}_{\epsilon}(\phi_c, D_c)$. % with respect to dataset $D_c$.
%
%In practice, there are different ways to inflate a bound and we may adjust it for the best behavior.
%
Finally, we propose using the constraint $\hat{U}_{\epsilon}(\phi_c, D_c) \le 0$ during candidate selection to find a candidate solution $\phi_c$ that is likely to pass the fairness test.

\subsubsection{Optimizing for a Candidate Solution With a Constrained Objective}\label{sec:objective}
In addition to finding a candidate solution that is likely to pass the fairness test, candidate selection also favors candidate solutions that have high expressiveness, so that the representations it generates are effective for downstream tasks.
We propose a candidate selection mechanism that achieves this goal without being limited to a specific learning algorithm. %In fact, our candidate selection is not limited to a specific design to fulfill this objective.
We support most parameterized representation learning architectures proposed in previous work, including the VAE-based methods~\citep{kingma2013autoencoding,Louizos2016Variational}, contrastive learning methods~\citep{Gupta2021Controllable,Oh2022Learning}, etc.
%
%Our framework can support any of these architectures.
%
In our experiments, we focus on an adaptation of VAE \citep{Louizos2016Variational} to construct the objective function that candidate selection optimizes.
%
%We note that we drop the Maximum Mean Discrepancy (MMD) regularization which encourages statistical independence between $S$ and $Z$ as proposed in VFAE, because it is an heuristic approach for encouraging fairness and becomes unnecessary under our proposed constraint.
%as it is one of the methods that does not require any labeled downstream task for training.
%
Specifically, we define $X \sim p_{\theta}(\cdot|Z,S)$ as the generative model for $X$ with input $(Z,S)$ parameterized by $\theta$.
Let $\mathbb{KL}$ denote KL-divergence, and $p(Z)$ be a  standard isotropic Gaussian prior, i.e., $p(Z) = \mathcal{N}(0, \mathbf{I})$, where $\mathbf{I}$ is the identity matrix.
Overall, we define the candidate selection process as approximating a solution to the constrained optimization problem:
\begin{align}\label{eq:objective}
\centering
    \max_{\theta, \phi}~~ & 
 \mathbb{E}_{q_\phi(Z|X,S)}\Big [\log p_\theta(X|Z,S)\Big ] - \mathbb{KL}\Big(q_\phi(Z|X,S) \| p(Z)\Big)\\
    \text{s.t. }~~ & \hat{U}_{\epsilon}(\phi, D_c) \le 0.
\end{align}

We propose using a gradient-based optimization to approximate an optimal solution $(\theta,\phi)$.
When using gradient based optimizers, the inequality constraint can be incorporated into the objective using the KKT conditions. %using gradient-based optimization with the KKT conditions. 
That is, we find saddle-points of the following Lagrangian function: 
\begin{equation}
\mathcal{L}(\theta,\phi; \lambda) \coloneqq -\mathbb{E}_{q_\phi(Z|X,S)}\Big[\log p_\theta(X|Z,S)\Big ] + \mathbb{KL}\Big(q_\phi(Z|X,S) \| p(Z)\Big) + \lambda \hat{U}_{\epsilon}(\phi, D_c),
\end{equation}
where $\lambda \ge 0$ is the Lagrange multiplier.

\vspace{-1.0mm}
\section{Theoretical Analysis}
\label{sec:theory}
\vspace{-1.0mm}
In this section we prove that %This section proves the following statement
\proj is a representation learning algorithm that provides the desired high confidence $\epsilon$-fairness guarantee, i.e., the probability that it produces a representation that is not $\epsilon$-fair for every downstream task and model is at most $\delta$. %any representation model it outputs has high probability to be $\epsilon$-fair, or equivalently, $\Delta_{\text{DP}}$ of arbitrary downstream model for any downstream task is upper-bounded by $\epsilon$.

We prove this claim in two steps. We first prove in Lemma~\ref{theorem:downstream} that if an algorithm $a$ satisfies $\Pr\left(\Tilde{g}_{\epsilon}(a(D)) \le 0 \right) \ge 1 - \delta$, then algorithm $a$ provides the $1-\delta$ confidence  $\epsilon$-fairness guarantee described in Def.~\ref{eq:seldonian}.
We then prove in Theorem \ref{theorem:conf} that \proj indeed satisfies $\Pr\left(\Tilde{g}_{\epsilon}(a(D)) \le 0 \right) \ge 1 - \delta$. Altogether, we can conclude that \proj guarantees with $1-\delta$ confidence that $\Delta_{\text{DP}}(\tau, a(D))$ is upper-bounded by $\epsilon$ for any $\tau$ (recall that here $a$ corresponds to \proj and $a(D)$ corresponds to the representation model parameters returned by \proj when run on dataset $D$).

\begin{lemma}\label{theorem:downstream}  If algorithm $a$ satisfies $\Pr\left(\Tilde{g}_{\epsilon}(a(D)) \le 0 \right) \ge 1 - \delta$, then algorithm $a$ provides the $1-\delta$ confidence  $\epsilon$-fairness guarantee described in Def.~\ref{eq:seldonian}. \textbf{\textit{Proof.}} 
See Appendix~\ref{apx:proof_lemma}.
\end{lemma}

\begin{theorem}\label{theorem:conf} \proj provides a $1-\delta$ confidence $\epsilon$-fairness guarantee. \textbf{\textit{Proof.}}
See Appendix~\ref{apx:proof_thm}.
\end{theorem}

% Prove why the chosen constraint and the adversarial learning can guarantee that any downstream task that uses the representation, is fair with a high confidence.

% \textbf{Proof: Suppose an oracle adversarial, then the representation learned has mutual information bounded by $\epsilon$ with high confidence}\\
% How does Seldonian optimization work?\\
% Why use adversarial learning?\\

% Why use information theory based approach? The other approach uses covariance distance which is hard to interpret?\\

% \textbf{Proof: Given a dataset, the optimal trivial solution of predicting sensitive attribute is the MLE or predicting the same class for all? Suppose MI is bounded by $\epsilon$, then the oracle predictor of sensitive attribute based on the representation is at best the trivial solution?}\\
% \textbf{Proof: no downstream predictor can do better than the adversarial at predicting sensitive attributes with the representations. Then it is fair wrt different metrics with high confidence.}

% tradeoff between utility and fairness? NSF?
% Should I start with a theoretical guarantee first, then introduce the algorithm and technical details?

% Personally, I feel that we should first introduce the notations, then introduce the seldonian algorithm, then prove the algorithm work theoretically, then discuss technical details.

% \textbf{Data processing inequality}
\vspace{-0.2cm}
\section{Practical Considerations for Upper Bounding $\Delta_\text{DP}$}
\label{sec:practical}

So far we have discussed using mutual information to upper bound $\Delta_\text{DP}$ (the violation of the demographic parity constraint), and ensure the $\epsilon$-fairness of a representation model with high confidence~(Sec.~\ref{sec:ft}).
Since $I(Z;S)$ is intractable, in Appendix~\ref{apx:upperbounds} we review four tractable upper bounds on $I(Z;S)$, and also discuss why in our experiments we adopt the first upper bound, $\Tilde{I}_1(Z;S)$, derived by~\citet[Section 2.2]{Song2019controllable}. We then test whether $\Tilde{I}_1(Z;S) \le \psi(\epsilon)$ to obtain the desired fairness guarantee (Eq.~\ref{eq:g_tilde}). %.
%
%We 

Because mutual information can be intractable, one might consider alternative methods for bounding $\Delta_\text{DP}$.
In Appendix~\ref{apx:alternative}, we explore potential alternatives for upper bounding $\Delta_\text{DP}$ but find limitations that prevent the adoption of these methods in \proj. % cannot adopt these methods.
%
%Given the limitations of these alternative approaches, 
Hence, we return to the original idea of using mutual information, $I(Z;S)$, to limit $\Delta_\text{DP}$.
However, using mutual information has another drawback we must overcome.
There tend to be significant gaps between $I(Z;S)$ and $\psi(\sup_{\tau}\Delta_\text{DP}(\tau, \phi))$, and between $\Tilde{I}_1(Z;S)$ and $I(Z;S)$ (demonstrated with an example in Appendix Fig.~\ref{fig:gap}). Hence, using the $\psi$-based bound on $\Tilde{I}_1(Z;S)$ results in exceedingly conservative bounds on $\Delta_\text{DP}$.
We 
analyze the gap between 
$I(Z;S)$ and $\psi(\sup_{\tau}\Delta_\text{DP}(\tau, \phi))$ in Appendix~\ref{apx:psigap}, and the gap between $\Tilde{I}_1(Z;S)$ and $I(Z;S)$ in Appendix~\ref{apx:migap}. Based on these analyses, one might consider using a constraint of the form \begin{align}\label{eq:adjustment}
    I(Z;S) \le \psi(\epsilon) + \gamma + \upsilon,
\end{align}
where $\gamma \ge 0$ and $\upsilon \ge 0$. If $\gamma$ lower bounds the gap between $I(Z;S)$ and $\phi(\epsilon)$, and $\upsilon$ lower bounds the gap between $\Tilde{I}_1(Z;S)$ and $I(Z;S)$, then $\Tilde{I}_1(Z;S)$ upper bounds $\psi(\Delta_\text{DP}(\tau, \phi))$.
One can then let \begin{align}
 \Tilde{g}'_{\epsilon}(\phi) \coloneqq \Tilde{I}_1(Z;S) - (\psi(\epsilon) + \gamma + \upsilon),
\end{align}
and use \proj to ensure $\Tilde{g}'_{\epsilon}(\phi) \le 0$, i.e., $\Tilde{I}_1(Z;S) \le \psi(\epsilon) + \gamma + \upsilon$, with high probability.

However, we do not know which values of $\gamma$ and $\upsilon$ satisfy $\gamma \le I(Z;S) - \psi(\epsilon)$ and $\upsilon \le \Tilde{I}_1(Z;S) - I(Z;S)$.
While one might consider treating $\gamma, \upsilon$ as tunable hyperparameters, doing so would compromise the algorithm's ability to provide a high-confidence guarantee that $\sup_\tau\Delta_\text{DP}(\tau, \phi) \le \epsilon$. 
 In summary, excluding $\gamma, \upsilon$ can make it difficult and sometimes impossible for any algorithm %\textcolor{red}{[Phil "our algorithm" sounds like this is a flaw with our method, when really it impacts all methods that might try to do this. Try to make this a more general statement.]}
 to return $\epsilon$-fair solutions with the desired confidence. % (that is, the algorithm frequently returns \texttt{NSF})
 However, including $\gamma, \upsilon$ (as hyperparameters) in our method prevents it from providing the high-confidence fairness guarantee as defined in Def.~\ref{def:highConfFairRep}.

We provide one way to select values for $\gamma$ and $\upsilon$ that likely lower bound the mutual information gaps, as detailed in Appendix.~\ref{apx:adjusted_frg}. Additionally, in Sec.~\ref{sec:exp_practical}, we present empirical evaluations  to demonstrate that \proj with these practical adjustments tends to satisfy $\Delta_\text{DP}(\tau, \phi) \le \epsilon$, although the formal guarantee is no longer ensured.
%
%\textcolor{red}{Appendix?} We note, however, that although this approach results in confidence intervals that hold with roughly the desired probability, it does not result in an actual high-confidence guarantee. Although this is not optimal, methods that tend to provide reasonable confidence intervals can often be useful even if the confidence intervals do not actually have guaranteed coverage (e.g., common applications of Student’s t-test to non-normal data and the use of bootstrap confidence interval~\citep{LearnedMiller2020new}).
%
%\textcolor{red}{[Phil: This paragraph below is important. There are some wording/grammar issues ,but the message is very good.]}

\vspace{-3mm}
\section{Experiments }\label{sec:exp}
\vspace{-1.5mm}
% \begin{figure}
% \vspace{-3.7mm}
% \includegraphics[trim={0.3cm 0.7cm 0.3cm 0.5cm},clip, width=0.95\textwidth]{iclr2023/Adult_main.pdf}
% \centering
% \vspace{-3.7mm}
% % \hspace{-3mm}
% \caption{\small{Comparisons between baselines and \proj with the practical adjustments (Sec.~\ref{sec:practical}) on \emph{Adult}.
% }}
% \label{fig:exp_adult}
% \vspace{-3.0mm}
% \end{figure}%\footnotemark\footnotetext{}
\textbf{Datasets.} We use 2 real-world datasets: UCI \emph{Adult}'s sensitive attribute is \emph{gender} and a downstream task predicts income;
%\footnote{\url{ https://archive.ics.uci.edu/ml/datasets/Adult}}
\emph{UTK-Face}'s sensitive attribute is \emph{ethnicity} and downstream labels are gender and age.
%\footnote{\url{https://susanqq.github.io/UTKFace/}}
We include detailed descriptions in Appendix~\ref{apx:datasets} and the statistics in Appendix Table~\ref{tab:dataset}.
%
%The third is the Heritage Health dataset~\TODO{link}, which contains information of over 60,000 patients. The downstream task is to predict whether the Charlson Index (an estimation of patient mortality) is greater than zero. We consider gender as the protected attribute as well.\\
%The first dataset is the UCI \textit{German} credit dataset\footnote{\url{https://archive.ics.uci.edu/ml/datasets/Statlog+(German+Credit+Data)}}. It contains
%information about 1000 individuals. The binary sensitive
%feature is whether the individual’s age exceeds
%a threshold. The downstream task is to predict whether the individual is offered credit or not.
%\textbf{Baselines.}\\

\textbf{Evaluations.} We are interested in addressing three research questions when evaluating \proj.
(1)~Do the empirical results align with our expectation that \proj produces $\epsilon$-fair representation models? In other words, is $\Delta_{\text{DP}}$ of all downstream models and tasks upper-bounded by a desired $\epsilon$ with high probability?
To address this question, we estimate the probabilities of violating the constraint $\Delta_{\text{DP}}(\tau, \phi) \le \epsilon$ using results from multiple runs of the algorithm with different training samples.
(2)~Can \proj learn expressive %representation models such that the 
representations that are useful for downstream predictions?
We evaluate the prediction performance on downstream tasks using the area
under the ROC curve (AUC).
(3)~Would \proj frequently result in \texttt{NSF} to avoid unfairness even when a sufficient amount of data and reasonable values of $\epsilon$ and $\delta$ are provided?
To address this question, we evaluate the probability that \proj provides a solution other than \texttt{NSF}.

\textbf{Experiment setup.} For each dataset, we hold out 20\% of the data as the \emph{test dataset}.
To assess the probabilities of $\Delta_\text{DP}$ being bounded by $\epsilon$ for all downstream models, we generate multiple training datasets from the remaining 80\% of the data and train multiple representation models on these datasets.
%To assess the probabilities of $\Delta_\text{DP}$ being bounded by $\epsilon$ for all downstream models, we must generate multiple training datasets.
To construct the training datasets, we initially generate 10 datasets by resampling 80\% of the data (excluding the test dataset) with replacement.
% of the with replacement from the remaining 80\% each having the same size as 80\% of the dataset. We achieve this by sampling with replacement from the remaining 80\% of the dataset, excluding the test dataset.
%
Next, we extract proportions of 10\%, 15\%, 25\%, 40\%, 65\% or 100\% from each resampled dataset above to create a single \emph{training dataset}.
In total, we apply each algorithm to 60 training datasets, resulting in 60 representation models. Subsequently, we assess the results using all these models on the test dataset.

For the remainder of this section, we start by providing an example of using FRG to find representation models with high-confidence fairness guarantees. 
We then apply the practical adjustments (as described in Sec.~\ref{sec:practical}) to \proj, and compare with competitive baselines. We conduct ablation studies on \proj with these adjustments in Appendix~\ref{apx:ablation}.\footnote{For Figures 3--4, each panel shows the mean (point) and standard deviation (shaded region) of a quantity as a function of the amount of data (in log scale). The black dashed line is set at the desired confidence level $\delta$.}

%show that \proj still tends to satisfy $\epsilon$-fairness more likely than baselines while maintaining comparable prediction performance, even though the guarantee no longer exists.
\vspace{-0.2cm}
\subsection{Evaluation on \proj That Provides High Confidence Fairness Guarantee}\label{sec:ideal}
\vspace{-0.13cm}
\begin{figure}
\centering
   \vspace{-6mm}

    % \includegraphics[trim={2.1cm 14.3cm 11.7cm 4.6cm},clip,width=0.5\textwidth]{TGAT-issue.pdf}
%trim=[left bottom right top]
\includegraphics[trim={1.0cm 0.9cm 0.6cm 0.76cm},clip,width=0.9999\textwidth]{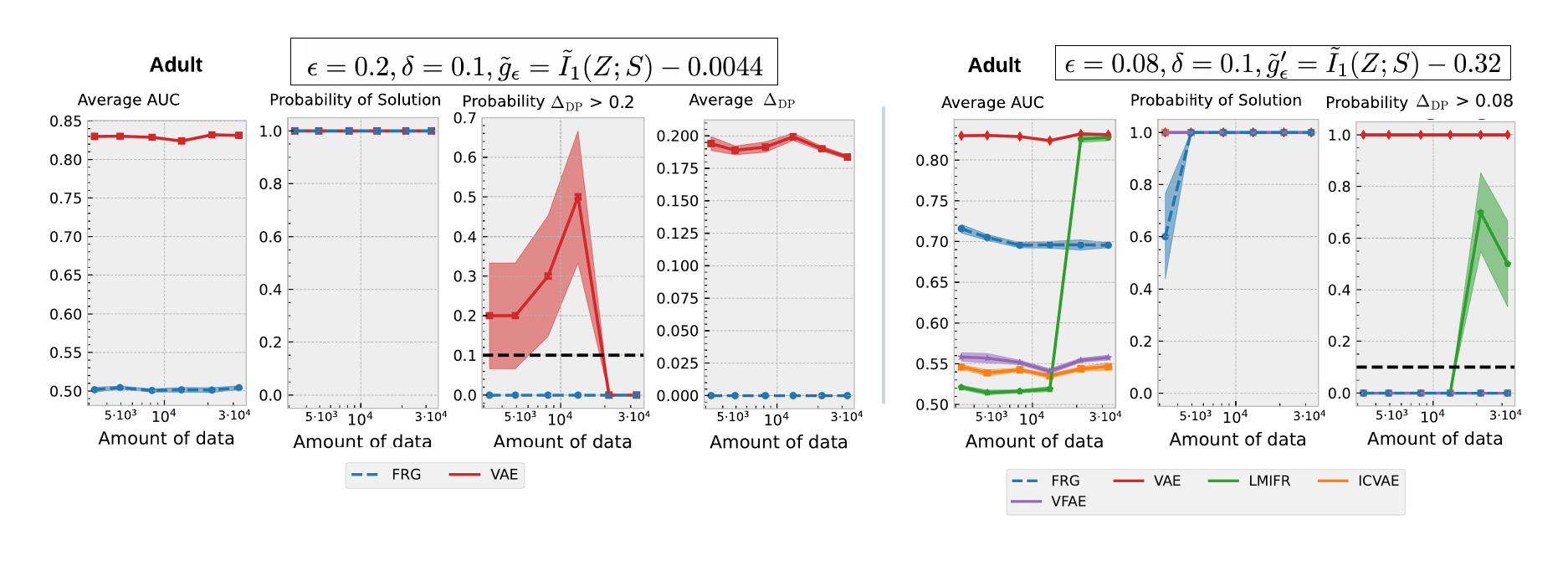}
%\hspace{-3mm}
\vspace{-7mm}
\caption{\small{On the \emph{left}, we give an example of employing FRG to provide high-confidence fairness guarantees (Def.~\ref{def:highConfFairRep}) on the \emph{Adult} dataset, including VAE as a baseline. On the \emph{right}, we show the comparisons between baselines and \proj with the practical adjustments (Sec.~\ref{sec:practical}) on \emph{Adult}. The first three plots on both sides are: (1) the average AUC when applying the representations to the designated downstream task, (2) the fraction of trials that returned a solution excluding \texttt{NSF}, and (3) the fraction of trials that violates $\Delta_{\text{DP}}(\tau, \phi) \le \epsilon$ on the ground truth dataset. The fourth plot on the left shows the average $\Delta_\text{DP}$ on the downstream task. % to facilitate discussions.
}}
\label{fig:exp_combine}
\end{figure}

We evaluate \proj that provides high-confidence fairness guarantees (Def.~\ref{def:highConfFairRep}) on the \emph{Adult} dataset.
For demonstration purposes, we select $\epsilon = 0.2$ and $\delta = 0.1$, which means that \proj guarantees with 90\% confidence that downstream models do not violate $\Delta_\text{DP}(\tau, \phi) \le 0.2$.
It is worth noting that the selected $\epsilon = 0.2$ is smaller than both the $\Delta_\text{DP}$ calculated with the true labels (as shown in Appendix Table~\ref{tab:dataset}), and the upper bound on $\Delta_\text{DP}$ calculated with the prediction labels from a predictor that achieves equalized odds~\citep[Theorem 3.1]{Zhao2020Conditional}.
We estimate $\Pr(S=1) \approx 0.668$ from the dataset, which yields $\psi(\epsilon) \approx 0.0044$.
We incorporate the constraint $\Tilde{I}_1(Z;S) \le \psi(\epsilon)$ to guarantee $\epsilon$-fairness with $1 - \delta$ confidence ($\Tilde{I}_1(Z;S)$ denotes the upper bound to $I(Z;S)$ as derived by~\citet[Section 2.2]{Song2019controllable}).
We include a vanilla VAE without any fairness consideration as a baseline.

We show the result in Fig.~\ref{fig:exp_combine} Left.
As demonstrated in the third and fourth plots, \proj violates the constraint $\Delta_\text{DP}(\tau, \phi) \le 0.2$ with a probability smaller than $0.1$, whereas VAE violates the constraint with a probability larger than $0.1$ when it uses less than 65\% of the training data.
According to the second plot, \proj can also return solutions (i.e., not \texttt{NSF}) for all the trials.

Nonetheless, as discussed in Sec.~\ref{sec:practical}, the constraint $\Tilde{I}_1(Z;S) \le \psi(\epsilon)$ is overly conservative, which leads to relatively low AUC on average, as illustrated in the first plot.
%The expressiveness of the representation models is also sacrificed and downstream prediction is under-performing, as shown by the first plot.
%
Additionally, the fourth plot demonstrates \proj's ability to consistently keep $\Delta_\text{DP}$ near zero.
Hence, applying an even stricter $\epsilon$ constraint on \proj for high-confidence fairness guarantees is impractical and unnecessary.

\vspace{-0.15cm}
\subsection{Evaluation on the Practical Adjustments on \proj}\label{sec:exp_practical}
\vspace{-1.mm}

\begin{figure}
\centering
\includegraphics[trim={0.6cm 0.7cm 0.6cm 0.84cm},clip,width=0.73\textwidth]{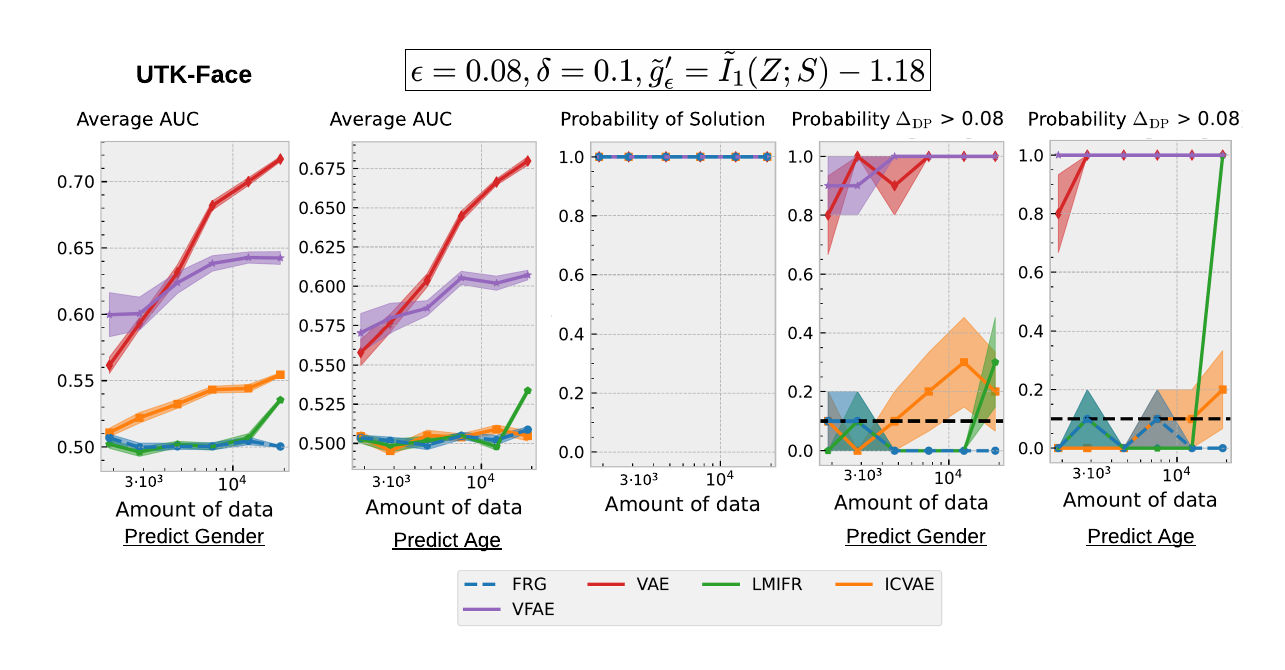}
\vspace{-3.7mm}
% \hspace{-3mm}
\caption{\small{
We show the comparisons between baselines and \proj with the practical adjustments (Sec.~\ref{sec:practical}) on \emph{UTK-Face}. Note that there are two downstream tasks, predicting gender and age. Since a learned representation model is used by all downstream tasks, there is only one probability of returning a solution excluding $\texttt{NSF}$. 
}}
\label{fig:exp_face}
% \vspace{-5.0mm}
\end{figure}

Considering the impracticality of strictly adhering to Def.~\ref{def:highConfFairRep} for \proj to provide high-confidence fairness guarantees, we evaluate \proj with the practical adjustments (as detailed in Sec.~\ref{sec:practical}) that guarantees the constraint $\Tilde{I}_1(Z;S) \le \psi(\epsilon) + \gamma + \upsilon$ is satisfied with high confidence.
We compare it with several baselines including LMIFR~\citep{Song2019controllable}, ICVAE~\citep{Moyer2018Invariant}, VFAE~\citep{Louizos2016Variational}and vanilla VAE. Detailed descriptions of these baselines are provided in Appendix~\ref{apx:baselines}.
%

%Given the impracticality of \proj that provides high-confidence fairness guarantees strictly following Def.~\ref{def:highConfFairRep}, we apply practical adjustments on \proj (Sec.~\ref{sec:practical}) which instead guarantees the constraint $\Tilde{I}_1(Z;S) \le \psi(\epsilon) + \gamma + \upsilon$ is satisfied with high confidence.
%
%We conduct further evaluations and compare with baselines.
%

We set $\epsilon=0.08, \delta = 0.1$ for evaluations on both the Adult and the UTK-Face datasets.
For \proj, We estimate $\phi(\epsilon) + \gamma + \upsilon$ according to Appendix~\ref{apx:adjusted_frg} and arrive at $\Tilde{g}'_\epsilon(\phi) = \Tilde{I}_1(Z;S) - 0.32$ for \emph{Adult}, and $\Tilde{g}'_\epsilon(\phi) = \Tilde{I}_1(Z;S) - 1.18$ for \emph{UTK-Face}. We note that as \proj does not restrict the choices of architectures for maximizing representation models' expressiveness and downstream prediction performance, we ensure fair comparisons by adopting the VAE-based primary objective as proposed in Sec.~\ref{sec:objective}, and a multilayer perceptron (MLP) with one hidden layer as the downstream model across all baselines.
We find hyperparameters that achieve $\Pr(\Delta_{\text{DP}}(\tau,\phi) > \epsilon) < \delta$ while maximizing AUC on one designated downstream task.
Detailed procedures for hyperparameter tuning are provided in Appendix~\ref{apx:hyperparam}.
We show the results for Adult in Fig.~\ref{fig:exp_combine} Right, and UTK-Face in Fig.~\ref{fig:exp_face}.

For both datasets, \proj can limit the probabilities that  $\Delta_{\text{DP}}(\tau,\phi) > 0.08$ to be at most $0.1$ while maintaining AUC values comparable to those of baselines.
\proj also exhibits a high probability of returning solutions, excluding \texttt{NSF}, in most cases.
The only exception occurs with the Adult dataset when data size is only 10\% of the total training data. 
Our hypothesis is that with a small amount of training data, it becomes challenging for candidate selection to recommend candidate solutions that satisfies $\Tilde{g}'_\epsilon(\phi) \le 0$ with high probability while simultaneously achieving high AUC.
%We hypothesize that with small amount of training data, it is hard for candidate selection to propose candidate solution that satisfies $\Tilde{g}'_\epsilon(\phi) \le 0$ with high probability while achieving high AUC.
%
%Given the hyperparameters we choose, candidate selection prioritize prediction performance.
%
%Fairness test computes the high-confidence upper bounds on $\Tilde{g}'_\epsilon(\phi)$ and rejects these candidate solutions because it does not have sufficient confidence that $\Tilde{g}'_\epsilon(\phi) \le 0$.

On the Adult dataset, among the methods that upper bound the probabilities of violating $\Delta_{\text{DP}}(\tau,\phi) \le 0.08$ by $1 - \delta$, \proj achieves the highest AUC.
Although LMIFR (evaluated with 65\% and 100\% of the training data) and VAE achieve higher AUC, they are more likely to violate the $\epsilon$-fairness constraint.
% with higher probability than the required confidence level $\delta$.
%
%For LMIFR evaluated with less training data, and for ICVAE and VFAE,  
In all other baseline evaluations, the need to tune hyperparameters to ensure high probabilities of $\Delta_{\text{DP}}(\tau,\phi) \le 0.08$ often results in subpar predictions in terms of AUC.

On UTK-Face, achieving $\epsilon$-fairness is more challenging with a multinomial and imbalanced sensitive attribute, especially when $\epsilon=0.08$.
Note that we tune hyperparameters and estimate $\Tilde{g}'_\epsilon(\phi)$ (Appendix~\ref{apx:adjusted_frg}) using only the gender labels while keeping the age labels hidden. The result shows that \proj can indeed effectively upper bound $\Delta_{\text{DP}}$ by a small $\epsilon$ across multiple downstream tasks with a high probability.
Although baseline methods can occasionally achieve higher AUC, and ICVAE and LMIFR can maintain $\Delta_\text{DP} \le 0.08$ most of the time, they cannot consistently ensure $\Delta_\text{DP}$ remains below 0.08 on either downstream task with sufficient probabilities.
This highlights the importance of providing high-confidence fairness guarantees when training fair representation models.

\vspace{-0.25cm}
\section{Related Work}\label{sec:rw}
\vspace{-0.2cm}
Fair representation learning (FRL) has been studied since at least 2013~\citep{Zemel2023Learning}.
%, when researchers studied methods for encoding as much information about input features as possible in vector representations, while limiting measures of the resulting unfairness when the representation is used for downstream prediction tasks 
 While some prior works \citep{Lahoti2019iFair, Lahoti2019Operationalizing, Peychev2022Latent,Ruoss2020Learning} focus on individual fairness,
 %(i.e., individuals with similar attributes should be treated similarly)
  we focus on group fairness
 %(i.e., similar predictive performance should be achieved across different groups) 
 with unfairness quantified by metrics like demographic parity, equalized odds, equal opportunity, and others~\citep{Dwork2012Fairness,Hardt2016Equality}. 

Several FRL studies prioritize optimization with respect to a specific downstream task~\citep{Calmon2017Optimized, McNamara2019Provably, zehlike2019matching, Calmon2018Data, Gordaliza2019Obtaining, Shui2019iFair, Zhu2021Learning, Rateike2022Dont}.
%to mitigate unfairness while retaining the utility of the representations 
%
In this paper, we instead focus on learning general representations that are fair, even when downstream tasks are unknown or unlabeled.

Numerous related studies aim to achieve FRL without relying on labels from downstream tasks~\citep{Hort2022Bias,Mehrabi2021Survey}.
%There are majorly two learning objectives, minimizing unfairness and maximizing utility. We discuss how related works achieve these two objectives separately.
%When we optimize for fairness representations, we cannot only consider fairness for some chosen downstream tasks, as a fair representation should be fair for any downstream task. 
%Most of these representation learning methods seek to mitigate unfairness, quantified by metrics like demographic parity, equalized odds, equal opportunity, and others~\citep{Dwork2012Fairness,Hardt2016Equality} across all downstream models and tasks by minimizing the inclusion of sensitive information in the learned representations.
%They accomplish this .
One category of these methods draws inspiration from information theory and probability theory, focusing on either reducing the mutual information between sensitive attributes and representations~\citep{ Song2019controllable, Jaiswal2020Invariant, Kairouz2021Generating, Kim2020Fair, Liu2022Fair, Moyer2018Invariant, Gupta2021Controllable}, or maximizing the conditional entropy of sensitive attributes given representations~\citep{Xie2017Controllable,Roy2019Mitigating,Sarhan2020Fairness}.
One work explores the use of distance covariance as an alternative to mutual information~\citep{Liu2022FairAlternative}.
Some methods can limit downstream unfairness by constraining the total variation distance between the representation distributions of different groups~\citep{Madras2018Learning, Zhao2020Conditional, Shen2021Fair,Balunovic2022Fair}.
Other approaches promote independence among sensitive attributes through penalization based on Maximum Mean Discrepancy~\citep{Louizos2016Variational,Oneto2020Exploiting,Deka2023MMD}, adversarial training that limits the adversary's performance in predicting sensitive attributes~\citep{Edwards2016Censoring, Feng2019Learning, Qi2021FairVFL,Wu2022Semi,Kim2022Learning}, meta-learning~\citep{Oneto2020Learning}, PCA~\citep{Fast2023Lee,Kleindessner2023Efficient}, measures for statistical dependence~\citep{Grari2021Learning,Quadrianto2019Discovering},
%such as Hirschfeld-Gebelein-R\'{e}nyi
learning a shared feature space between groups~\citep{Cerrato2021Fair}, or disentanglement~\citep{Creager2019Flexibly, Oh2022Learning, Locatello2019On}.

Some FRLs provide theoretical analyses. \citet{Madras2018Learning, Gupta2021Controllable, Zhao2020Conditional,Shen2021Fair}  offer provable upper bounds on the unfairness of all downstream models and tasks. %Their algorithms are designed to encourage these upper bounds to be small, thereby limiting unfairness.
\citet{Gitiaux2021Learning} present a method to compute an empirical upper bound on the expected values of unfairness
%(measured by demographic parity)
for all downstream models and tasks.
Some recent work~\citep{Jovanovic2023FARE, Balunovic2022Fair} provides practical certificates that serve as high-confidence upper bounds on the unfairness,
%(measured by demographic parity)
 using finite samples, but their methods are limited to particular representation model architectures (e.g., decision trees, normalizing flows) and representation distributions (e.g., discrete), which may not be ideal in some situations.
Furthermore, these methods do not provide a framework for learning fair representation models with high confidence. Specifically, they do not accept a user-defined threshold and ensure with high confidence that the unfairness of all downstream models is upper-bounded by that threshold.

Finally, some prior work~\citep{Li2022Fairee, Philip2019Preventing,Hoag2023Seldonian} provides high-confidence guarantees for fair classification, but does not explore the representation learning setting.
\vspace{-0.22cm}
\section{Conclusion and Future Work}
\vspace{-0.15cm}
In this work, we introduced \proj, a fair representation learning framework that provides high-confidence fairness guarantees, ensuring that unfairness for all downstream models and tasks is upper-bounded by a user-defined threshold. After substantiating our work with theoretical analysis, we conducted empirical evaluations that demonstrate \proj's effectiveness in upper-bounding unfairness across various downstream models and tasks. In the future, we plan to extend \proj for representation learning with other guarantees related to privacy, safety, robustness, and more.
% In this work, we introduced \proj, a fair representation learning framework that provides high-confidence fairness guarantees, ensuring that the unfairness for all downstream models and tasks is upper-bounded by a user-defined threshold.
% %\proj consists of two key components: \emph{candidate selection} and \emph{fairness test}, which collaborate to generate representation models that are fair with high probabilities while maintaining high expressiveness.
% After substantiating our work with theoretical analysis, we conducted empirical evaluations that demonstrates \proj's effectiveness in upper bounding the unfairness across various downstream models and tasks. In the future, we plan to extend \proj for representation learning with other guarantees that involve privacy, safety, invariance and etc.
\section{Reproducibility Statement}
We provide the source code in an repository \href{https://github.com/JamesLuoyh/FRG}{here}.\footnote{\url{https://github.com/JamesLuoyh/FRG}} The detailed instructions for creating the environment, acquiring the datasets, and running \proj and baselines are presented in \texttt{README.md}. One may refer to Section~\ref{sec:exp} to get the details of experiment setup and may refer to Appendix~\ref{apx:datasets},~\ref{apx:baselines},~\ref{apx:hyperparam} to get the details of the datasets, the baselines and the hyperparameter tuning procedure.
\section{Acknowledgements}
This work is supported by the National Science Foundation under grant no. CCF-2018372 and by a gift from the Berkeley Existential Risk Initiative.
\bibliography{iclr2024_conference}
\bibliographystyle{iclr2024_conference}

\appendix
\newpage
\appendix

\section{Details of Property~\ref{eq:dpbound}}
\citet{Gupta2021Controllable} has derived Property~\ref{eq:dpbound} where $I(Z;S)$ is an upperbound for a strictly increasing non-negative convex function in $\Delta_{\text{DP}}$ of any $\tau$, which we denote as $\psi$.
\citet{Gupta2021Controllable} has also found that when $I(Z;S) = 0$, $\psi(\Delta_{\text{DP}}(\tau, \phi)) = 0$ and $\Delta_{\text{DP}}(\tau, \phi) = 0$.
We now define $\psi$ in detail by first introducing a helper function $f$.\\

\begin{definition}[A helper function $f$]\label{apx:f}
$$f(V) = \max\left(\log\left(\frac{2 + V}{2-V}\right) - \frac{2V}{2+V}, \frac{V^2}{2} + \frac{V^4}{36} + \frac{V^6}{288}\right).
$$ with domain $V\in[0,2)$.
    
\end{definition}

\begin{definition}[function $\psi$ with parameter $\Delta_{\text{DP}}(\tau, \phi)$]\label{apx:psi} When $S$ is binary, and $f$ follows Def.~\ref{apx:f}, 
$$\psi(\Delta_{\text{DP}}(\tau, \phi)) = (1-\pi)f(\pi\Delta_{\text{DP}}(\tau, \phi)) + \pi f((1-\pi)\Delta_{\text{DP}}(\tau, \phi))$$
where $\pi = P_s(S=1)$ with $P_s$ as the marginal distribution of $S \in \{0,1\}$.

When $S$ is multinomial with $K$ classes,
$$\psi(\Delta_{\text{DP}}(\tau, \phi)) = f(\alpha\Delta_{\text{DP}}(\tau, \phi)), \alpha = \min_{k = 1, \ldots, K}\pi_k,$$ where $\pi_k = P_s(S=k)$ with $P_s$ as the marginal distribution of $S \in \{1, \ldots, K\}$.

\end{definition}
\section{Proof of Lemma~\ref{theorem:downstream}}\label{apx:proof_lemma}

\begin{lemma}[Lemma~\ref{theorem:downstream} restated] If algorithm $a$ satisfies $\Pr\left(\Tilde{g}_{\epsilon}(a(D)) \le 0 \right) \ge 1 - \delta$, then algorithm $a$ provides the $1-\delta$ confidence  $\epsilon$-fairness guarantee described in Def.~\ref{eq:seldonian}.
\end{lemma}

\textit{Proof.} 
Suppose $\Pr\left(\Tilde{g}_{\epsilon}(a(D)) \le 0 \right) \ge 1 - \delta$. By Eq.~\ref{eq:g_tilde}, $\Tilde{g}_{\epsilon}(a(D)) = \Tilde{I}(Z;S) - \psi(\epsilon) \ge I(Z;S) - \psi(\epsilon)$.
By property \ref{eq:dpbound}, $I(Z;S) \ge \sup_{\tau} \psi(\Delta_\text{DP}(\tau, a(D)))$.
So, the event $\left(\Tilde{g}_{\epsilon}(a(D)) \le 0 \right)$ implies that$\left(I(Z;S) - \psi(\epsilon) \le 0 \right)$, which further implies $\left(\sup_{\tau} \psi(\Delta_\text{DP}(\tau, a(D))) - \psi(\epsilon) \le 0 \right)$.
Using the fact that $\psi$ is strictly increasing in $[0,1]$ (Appendix \ref{apx:psi}), we have the following equivalent events:
\begin{align}
    &\left(\sup_{\tau} \psi(\Delta_\text{DP}(\tau, a(D))) - \psi(\epsilon) \le 0 \right) \\
    \iff & \left(\psi(\sup_{\tau} \Delta_\text{DP}(\tau, a(D))) \le \psi(\epsilon) \right) \\
    \iff & \left(\sup_{\tau} \Delta_\text{DP}(\tau, a(D)) \le \epsilon \right) \\
    \iff & \left(\sup_{\tau} \Delta_\text{DP}(\tau, a(D)) - \epsilon \le 0 \right)\\
    \iff & \left( g_{\epsilon}(a(D)) \le 0 \right).
\end{align}
It follows that $\Pr\left( g_{\epsilon}(a(D)) \le 0 \right) \ge \Pr \left(\Tilde{g}_{\epsilon}(a(D)) \le 0 \right) \ge 1 - \delta$.
So, \proj (algorithm $a$) provides the desired $1-\delta$ confidence $\epsilon$-fairness guarantee described in Def.~\ref{eq:seldonian}, completing the proof.

\section{Proof of Theorem~\ref{theorem:conf}}\label{apx:proof_thm}
\begin{theorem}[Theorem~\ref{theorem:conf} restated] \proj provides the $1-\delta$ confidence $\epsilon$-fairness guarantee described in Def.~\ref{eq:seldonian}.
\end{theorem}

\textit{Proof.}
By Lemma \ref{theorem:downstream}, if \proj satisfies $\Pr\left(\Tilde{g}_{\epsilon}(a(D)) \le 0 \right) \ge 1 - \delta$, then \proj provides the desired $1-\delta$ confidence $\epsilon$-fairness guarantee. We prove by contradiction that when $a$ represents \proj, $\Pr\left(\Tilde{g}_{\epsilon}(a(D)) \le 0 \right) \ge 1 - \delta$.

We begin by assuming the result is false and then derive a contradiction. The beginning assumption is that 
$\Pr\left(\Tilde{g}_{\epsilon}(a(D)) \le 0 \right) < 1 - \delta$.
By contrapositive, we have 
$\Pr\left(\Tilde{g}_{\epsilon}(a(D)) > 0 \right) \ge \delta$.
By the construction of \proj, $a(D)$ is either \texttt{NSF} or the proposed candidate solution $\phi_c \in \Phi$. Notice that $\Tilde{g}_{\epsilon}(a(D)) > 0$ if and only if $a(D)$ does not return \texttt{NSF} but returns $\phi_c$ instead, i.e., $a(D)=\phi_c$.
The fairness test in \proj returns $\phi_c$ if and only if $U_{\epsilon}(\phi_c, D_f) \le 0$ (Sec.~\ref{sec:ft}).
Therefore, the following events are equivalent ($\Pr(A,B)$ denotes the joint probability of $A$ and $B$):
\begin{align}
    &(\Tilde{g}_{\epsilon}(a(D)) > 0) \\
    \iff & (\Tilde{g}_{\epsilon}(a(D)) > 0, a(D) = \phi_c, U_{\epsilon}(\phi_c, D_f) \le 0) \\
    \iff & (\Tilde{g}_{\epsilon}(\phi_c) > U_{\epsilon}(\phi_c, D_f), a(D) = \phi_c).
\end{align}
The joint event $(\Tilde{g}_{\epsilon}(\phi_c) > U_{\epsilon}(\phi_c, D_f), a(D) = \phi_c)$ implies $(\Tilde{g}_{\epsilon}(\phi_c) > U_{\epsilon}(\phi_c, D_f))$.
Therefore, $$\Pr\left(\Tilde{g}_{\epsilon}(\phi_c) \ > U_{\epsilon}(\phi_c, D_f)\right) \ge \Pr\left(\Tilde{g}_{\epsilon}(\phi_c) > U_{\epsilon}(\phi_c, D_f), a(D) = \phi_c\right) \ge \delta.$$

However, by construction of the fairness test, $\Pr\left(\Tilde{g}_{\epsilon}(\phi_c) \le U_{\epsilon}(\phi_c, D_f) \right) \ge 1 - \delta$ (Inequality~\ref{eq:confidence}), which implies $\Pr\left(\Tilde{g}_{\epsilon}(\phi_c) > U_{\epsilon}(\phi_c, D_f) \right) < \delta.$
This gives a contradiction, completing the proof.

 We note that this theorem is true for any choice of candidate selection, as the proof assumes the candidate solution $\phi_c$ is arbitrary.

\section{The Tractable Upper Bounds on $I(Z;S)$}\label{apx:upperbounds}
To our best knowledge, there are four tractable upper bounds on mutual information $I(Z;S)$ as derived by previous work~\citep{Song2019controllable,Moyer2018Invariant, Gupta2021Controllable}.
Next, we discuss these approaches and their limitations. Although our general approach is compatible with any upper bound on mutual information, given the limitations of each method, we consider the first of the two approaches ($\Tilde{I}_1(Z;S)$ below) by \citet{Song2019controllable} the most suitable in practice. Thus, we only adopt $\Tilde{I}_1(Z;S)$ in our experiments. 
%In our experiments, we follow the adversarial training approach proposed by~\citet{Song2019controllable} and we argue that it is the only practical upper bound for learning fair representation models with high confidence. We state the caveats of each of the four upper bounds in the following paragraphs. 

\citet{Song2019controllable}
proposed two upper bounds on $I(Z;S)$.

\textbf{$\Tilde{I}_1(Z;S)$: the first upper bound derived by \citep[Section 2.2]{Song2019controllable}.}
We denote the first upper bound as $\Tilde{I}_1(Z;S)$ and $\Tilde{I}_1(Z;S) \ge I(Z;X,S) \ge I(Z;S)$~\citep[Section 2.2]{Song2019controllable}.
This is a theoretically guaranteed upper bound. We discuss the limitation of this upper bound in Appendix~\ref{apx:migap} that using this upper bound may diminish the expressiveness of the representations. However, we still find it effective for \proj to limit $\Delta_{\text{DP}}$ by $\epsilon$ in experiment (Sec.~\ref{sec:exp}).
%, and it is necessary to introduce another hyperparameter $\upsilon$ to estimate the gap between $I(Z;X,S)$ and $I(Z;S)$. % be loose and hard to be kept small as well.
%

%
%It makes it difficult, and sometimes impossible to find a solution that satisfies the constraint with high confidence. Even if we find a solution, the low $I(X;Z|S)$ implies that $Z$ is not an expressive representation that gives enough utility for downstream models.

%\textcolor{blue}{Why does the above imply the following sentence? Explain little more to the reader.}
%This makes the upper bound impractical because it either diminishes the expressiveness of the representation model significantly or become too loose to have a solution.

\textbf{$\Tilde{I}_2(Z;S)$: the second upper bound derived by \citep[Section 2.3]{Song2019controllable}.} \citet{Song2019controllable} proposed a tighter upper bound compared to $\Tilde{I}_1(Z;S)$, which we denote as $\Tilde{I}_2(Z;S)$. However, it requires adversarial training, and the true upper bound can only be obtained when the adversarial model approaches global optimality.
This is not ideal because if the adversarial model is under-performing, we may under-estimate the upper bound to $I(Z;S)$, and guaranteeing $\Tilde{I}_2(Z;S) \le \psi(\epsilon)$ does not guarantee $I(Z;S) \le \psi(\epsilon)$ or $\epsilon$-fairness. This result has also been confirmed by prior work including~\citet{Xu2021Theory,Elazar2018Adversarial,Gupta2021Controllable} and~\citet{Gitiaux2021Learning}.
%Although the high confidence fairness guarantee may be compromised when the adversarial model does not achieve optimum, %\textcolor{red}{[Do we really know this is the only reasonable tight upper bound? Could there be one we didn't see in some other paper? Could there be one that nobody has discovered yet? This claim is too strong.]}
%it is a reasonably tight upper bound that can be estimated unbiasedly with samples from a dataset, and so it is suitable for enforcing the constraint using this bound. In experiment, we observe that \proj using this upper bound sometimes under estimates $I(Z;S)$ and the constraint $\Delta_{\text{DP}} \le \epsilon$ is violated.

\textbf{$\Tilde{I}_3(Z;S)$: the upper bound derived by \citet{Moyer2018Invariant}.} \citet{Moyer2018Invariant} found $I(Z;S) = I(Z;X) - H(X|S) + H(X|Z,S)$ where $H$ denotes entropy.
They proposed ignoring the unknown positive constant term $H(X|S)$ and using the reconstruction error, i.e., $- \mathbb{E}_{q_\phi(Z|X,S)}\Big [\log p_\theta(X|Z,S)\Big ]$ to be an upper bound of $H(X|Z,S)$~\citep[Equations 2--7]{Moyer2018Invariant}.
Let $\Tilde{I}_3(Z;S) \coloneqq I(Z;X) - \mathbb{E}_{q_\phi(Z|X,S)}\Big [\log p_\theta(X|Z,S)\Big ]$.
Suppose $\Tilde{I}_3(Z;S) - I(Z;S) > \psi(\epsilon) + \gamma$ ($\gamma$ is defined in Inequality~\ref{eq:adjustment}), then any $\phi$ that satisfies $\Tilde{I}_3(Z;S) \le \psi(\epsilon) + \gamma$ would result in  $I(Z;S) \le 0$, which is a constraint that is impossible to satisfy.
%
% One may introduce another hyperparameter $\gamma_b \ge 0$ and adopt a constraint $\Tilde{I}_2(Z;S) \le \psi(\epsilon) + \gamma + \gamma_b$ so that $\Delta_\text{DP}$ can be bounded by $\epsilon$ if $\gamma_b$ lower bounds $\Tilde{I}_2(Z;S) - I(Z;S)$.
% %
Moreover, it can be difficult to estimate the gap $\Tilde{I}_3(Z;S) - I(Z;S)$ because (1) $H(X|S)$ is hard to estimate; (2) $\Tilde{I}_3(Z;S)$ is sensitive to the performance of the reconstruction model. 
%such that $H(X|S) - \gamma_H \le \psi(\epsilon) + \gamma$, while $\gamma_2 \le H(X|S)$, especially when $X$ is high-dimensional and $H(X|S)$ is large.
%\textcolor{blue}{TODO: Replace $\gamma$ here with a different symbol like $\upsilon$ since we already used $\gamma$ above.}
% This could be resolved by making $\gamma$ larger, but it is not clear how large $\gamma$ should be
% We could make the $\gamma$ larger but we find it less controllable as $H(X|S)$ can be in a much larger scale.

\textbf{$\Tilde{I}_4(Z;S)$: the upper bound derived by \citet{Gupta2021Controllable}.} \citet{Gupta2021Controllable} observed that $I(Z;S) = I(Z;S|X) + I(Z;X) - I(Z;X|S)$. They then derived a lower bound for the term $I(Z;X|S)$ using constrative estimation so that $I(Z;S)$ can be upper-bounded. %, they proposed to lower bound it via contrastive estimation.
Specifically, they proved $I(Z;X|S) \ge \mathbb{E}_{p(X,Z,S)}\big[\log \frac{e^{f(X,Z,S)}}{\frac{1}{M}\sum^M_{m=1}e^{f(X_m,Z,S)} }\big]$, where $p(X,Z,S)$ is the joint distribution of $(X,Z,S)$, $X_{1}, \cdots,X_{M} \sim p_{X|S}$, $p_{X|S}$ is the conditional distribution of $X$ given $S$, and $f$ is an arbitrary function \citep[Proposition 5]{Gupta2021Controllable}.
Since the distribution $p_{X|S}$ is unknown, the authors use the $X, S$ pairs in the dataset as samples from this conditional distribution.
When making a point estimate of the expectation, they use one sample from the dataset to evaluate the numerator, and use $M$ samples from the same dataset to evaluate the denominator.
This means that the estimation of the expectation can be biased because the point estimates are not independent. %each of the point estimates depend on each other.
Empirically, we also observe this issue and find that it tends to result in over-estimates of $I(Z;X|S)$ and under-estimates of $I(Z;S)$. Given how these terms are used in the expression for $I(Z;S)$, this results in bounds on mutual information that do not hold. %(\TODO{See appendix for more details}).

\section{Alternative Methods for Upper-bounding $\Delta_\text{DP}$}\label{apx:alternative}
%So far we have discussed using mutual information to upper bound $\Delta_\text{DP}$ (the violation of the demographic parity constraint). 
%
One might consider alternative methods for bounding $\Delta_\text{DP}$ because mutual information can be intractable and there can be a significant gap between mutual information and $\psi(\Delta_\text{DP})$ (that is, the upper bound can be loose). 
Several alternative methods have been proposed, which can provide bounds on $\Delta_\text{DP}$ using bounds on the total variation between the conditional distributions $p_{\tau, \phi}(\hat{Y}|S = 0)$ and $p_{\tau,\phi}(\hat{Y}|S = 1)$~\citep{Zhao2020Conditional,Madras2018Learning, Shen2021Fair,Balunovic2022Fair}. 
However, to our knowledge, there is not a known function such as $\psi$ (Appendix \ref{apx:psi}) that expresses the relation between the total variation and demographic parity, so total variation cannot be used to upper bound $\sup_{\tau}\Delta_\text{DP}(\tau, \phi)$ with a specific $\epsilon$. 
In other work, \citet[Section 5]{Jovanovic2023FARE} proposed a practical certificate that upper bounds $\sup_{\tau}\Delta_\text{DP}(\tau, \phi)$. 
However, their method requires $Z$ to be a discrete random variable, which is restrictive for general representation learning. 
Therefore, these methods are not suitable for our framework as they cannot be used to learn $\epsilon$-fair representation models with a high-confidence guarantee.

\begin{figure}
\centering
   % \vspace{-3mm}

    % \includegraphics[trim={2.1cm 14.3cm 11.7cm 4.6cm},clip,width=0.5\textwidth]{TGAT-issue.pdf}
\includegraphics[trim={0.2cm 0.1cm 0.3cm 0.4cm},clip,width=0.75\textwidth]{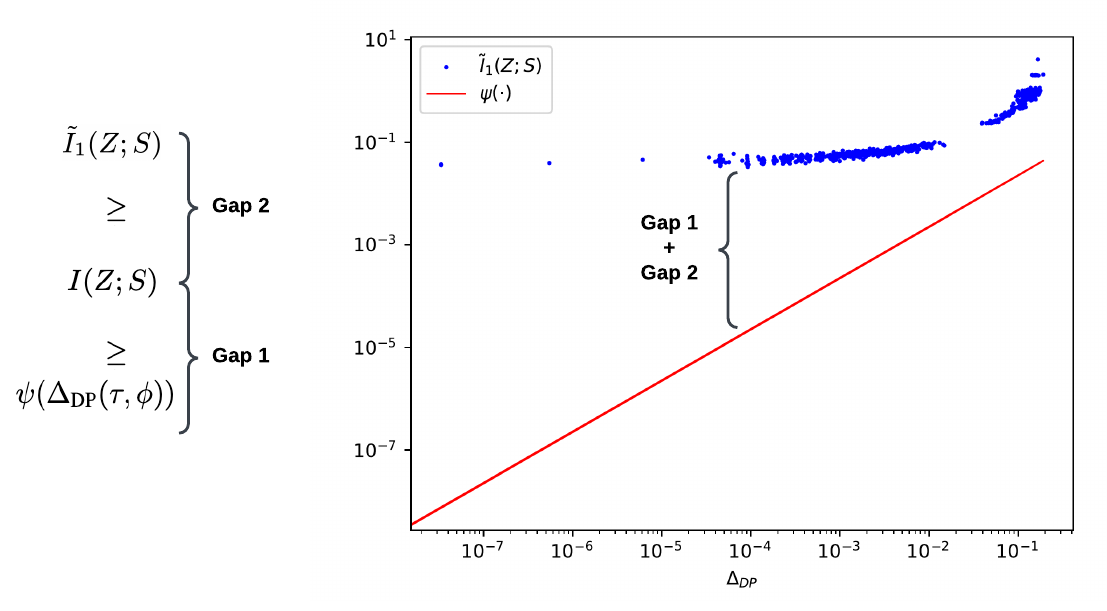}
%\hspace{-3mm}
% \vspace{-5mm}
\caption{\small{Using the \emph{Adult} dataset (details in Appendix.~\ref{apx:datasets}), we run L-MIFR~\citep{Song2019controllable} with different hyper-parameters to find representation models that achieve different $\Delta_{\text{DP}}(\tau, \phi)$. For each of the representation model, we record the corresponding tractable upper bound to $I(Z;S)$ by~\citet[Section 2.2]{Song2019controllable}, denoted as $\Tilde{I}_1(Z;S)$, and make the scatter plot in blue. We plot the function $\psi(\cdot)$ (Appendix~\ref{apx:psi}) in red. We highlight that there exists a gap between  $\Tilde{I}_1(Z;S)$ and $\psi(\Delta_{\text{DP}}(\tau, \phi))$, which consists of two gaps, $\Tilde{I}_1(Z;S) - I(Z;S)$ and $I(Z;S) - \psi(\Delta_{\text{DP}}(\tau, \phi))$, and it can be observed empirically as shown by the plot.}}
\label{fig:gap}
\end{figure}

\section{The Non-trivial Gap between $I(Z;S)$ and $\psi(\sup_{\tau}\Delta_\text{DP}(\tau, \phi))$}\label{apx:psigap}
%Given the limitations of the alternative approaches, we return to the original idea of using mutual information, $I(Z;S)$, to limit $\Delta_\text{DP}$, and ensure the $\epsilon$-fairness of a representation model with high confidence as described in Sec.~\ref{sec:ft}.
%
%, we evaluate whether $\Tilde{I}(Z;S) \le \psi(\epsilon)$ with high confidence to determine $\epsilon$-fairness of a representation model.
%
%However, using mutual information also has drawbacks that we must overcome. 
%However, it is necessary for us to consider the practicality of this evaluation.
%
In this section, we analyze the non-trivial gap between $I(Z;S)$ and $\psi(\sup_{\tau}\Delta_\text{DP}(\tau, \phi))$ that makes it difficult for any algorithm to obtain $\epsilon$-fairness.

As shown by~\citet[Figure 6]{Gupta2021Controllable}, there tends to be a significant gap between $I(Z;S)$ and $\psi(\sup_{\tau}\Delta_\text{DP}(\tau, \phi))$. Using their Figure 6 as an example, when $I(Z;S) \approx 0.035$, $\Delta_\text{DP}(\tau, \phi) \approx 0.15$ and $\psi(\Delta_\text{DP}(\tau, \phi)) \approx 0.0025$.
So, to ensure that $\Delta_\text{DP}(\tau, \phi) \le 0.15$ with high confidence using the $\psi$-based bound on mutual information, one must ensure that $I(Z;S) \le 0.0025$ with high confidence. 
However, in reality ensuring that $\Delta_\text{DP}(\tau, \phi) \le 0.15$ only requires $I(Z;S) \le 0.035$.
Obtaining a solution that satisfies $I(Z;S) \le 0.0025$ is far more difficult than obtaining a solution that satisfies $I(Z;S) \le 0.035$, and hence using the $\psi$-based bound on mutual information results in exceedingly conservative bounds on $\Delta_\text{DP}$. % It would be much harder to get a solution that satisfies $I(Z;S) \le 0.0025$ and it is usually an overkill for $\Delta_\text{DP}(\tau, \phi)$.

\section{The Non-trivial Gap between $\Tilde{I}_1(Z;S)$ and $I(Z;S)$}\label{apx:migap}
In this section we analyze the non-trival gap between $\Tilde{I}_1(Z;S)$ and $I(Z;S)$ where $\Tilde{I}_1(Z;S)$~(Appendix~\ref{apx:upperbounds}) is one of the upper bounds to $I(Z;S)$ as derived by~\citet[Section 2.2]{Song2019controllable}. We begin by analyzing the gap between $I(Z;X,S)$ and $I(Z;S)$. % to this upper bound.
$I(Z;X,S) - I(Z;S) 
%= H(Z) - H(Z|S) - (H(Z) - H(Z|X,S)) 
= H(Z|S) - H(Z|X,S) 
%= H(Z,S) - H(S) - H(Z,S,S) + H(X,S) 
= H(X|S) - H(X|Z,S) = I(X;Z|S)$.
This is the mutual information between $X$ and $Z$ given $S$, which is closely related to the primary objective we hope to maximize.
Overall, we have the following:%\textcolor{red}{[Phil: I put in an align block for now. We can tighten up for space when needed at the end]}
\begin{align}
    I(Z;S) \le& I(Z;X,S) \\
    =& I(Z;S) +  I(X;Z|S)\\
    \le& \Tilde{I}_1(Z;S)
\end{align}
%
% One may consider introducing a hyperparameter $\upsilon \ge 0$ and adopting the constraint 
% \begin{align}\label{eq:i1_constraint}
%     \Tilde{I}_1(Z;S) \le \psi(\epsilon) + \gamma + \upsilon,
% \end{align} as discussed in Sec.~\ref{sec:practical}, which ensures that $\Delta_\text{DP}$ is upper-bounded by $\epsilon$ if $\upsilon\le \Tilde{I}_1(Z;S) - I(Z;S)$ and $\gamma \le I(Z;S) - \psi(\epsilon)$.
% that ensures Inequality~\ref{eq:adjustment} is satisfied if $\upsilon\le \Tilde{I}_1(Z;S) - I(Z;S)$, and that $\Delta_\text{DP}$ is bounded by $\epsilon$ if $\gamma \le I(Z;S) - \psi(\epsilon)$.
%
Although using a constraint $\Tilde{I}_1(Z;S) \le \psi(\epsilon)$ encourages both $I(Z;S)$ and $I(X;Z|S)$ to be small which seems to diminish the expressiveness of the representation model, we show empirically that it is effective for upper bounding mutual information and the $\Delta_{\text{DP}}$ of the downstream tasks with high probability in experiment (Sec.~\ref{sec:exp}).% We discuss how we apply this updated constraint in \proj in detail in Appendix~\ref{apx:adjusted_frg}.
% \textcolor{red}{[Phil: Say whether we will use this approach or not in our experiments, and hint at how we set $gamma$.]}

%
%However, we do not find a way to either quantify or remove this gap and it is a necessary adjustment in practice so that the algorithm can find and output solutions.

\section{Practical Adjustments on \proj}\label{apx:adjusted_frg}

 In this section, we detail an approach to construct the practical constraint $\Tilde{I}_1(Z;S) \le \psi(\epsilon) + \gamma + \upsilon$ (Inequality~\ref{eq:adjustment}), and apply the constraint on \proj so that $\Delta_\text{DP}$ is likely to be bounded by $\epsilon$. We note, however, that although this approach results in confidence intervals that hold with roughly the desired probability, it does not result in an actual high-confidence guarantee. Although this is not optimal, methods that tend to provide reasonable confidence intervals can often be useful even if the confidence intervals do not actually have guaranteed coverage (see, for example, common applications of Student's $t$-test to non-normal data and the use of bootstrap confidence intervals~\citep{LearnedMiller2020new}). %\textcolor{red}{[Phil: I added text here. I think we need to talk about this a bit more, as with the added text. A reference would help - perhaps this one, which talks about t-test not holding? \url{https://arxiv.org/pdf/1905.06208.pdf} A reference that isn't by me would be stronger though.]}%, although there is no guarantee.%to find representation models that satisfy the practical constraint with high confidence.
 %

 % We first visualize in Fig.~\ref{fig:gap} and confirm that the gap between $\Tilde{I}_1(Z;S)$ and $\psi(\Delta_\text{DP}(\tau, \phi))$ indeed exists empirically.
 % %
 % The $\Tilde{I}_1(Z;S)$ and $\Delta_\text{DP}$ are recorded when we run the algorithm L-MIFR proposed by~\citet{Song2019controllable}  for multiple times on the Adult dataset (details in Sec.~\ref{sec:exp}).
 % %
 To construct the constraint 
$\Tilde{I}_1(Z;S) \le \psi(\epsilon) + \gamma + \upsilon$, we do not need to determine $\gamma$ and $\upsilon$ separately.
We only need to determine $\gamma + \upsilon$, and we want $\gamma + \upsilon$ to under-estimate the true gap $\Tilde{I}_1(Z;S) - \sup_\tau\psi(\Delta_\text{DP}(\tau, \phi))$ so that satisfying $\Tilde{I}_1(Z;S) \le \psi(\epsilon) + \gamma + \upsilon$ implies $\sup_\tau\psi(\Delta_\text{DP}(\tau, \phi)) \le \psi(\epsilon)$, and the representation model $\phi$ is $\epsilon$-fair.
%$\textcolor{red}{[Phil: Once you define $\epsilon$-fair earlier, you don't need to put it in quotes when you use it later, e.g., here.]}.
%
We propose a practical way of estimating a value for $(\gamma + \upsilon)$ using L-MIFR as follows.
(1)~Utilize the candidate selection data $D_c$ to run L-MIFR various hyperparameter settings, aiming to achieve $ \epsilon - c \le \Delta_\text{DP}(\tau, \phi) \le \epsilon$ on $D_c$, where $ 0 \le c \le \epsilon$ is a predetermined hyperparameter. For each value of $\Delta_\text{DP}(\tau, \phi)$, record the corresponding $\Tilde{I}_1(Z;S)$ on $D_c$.
(2)~Arrange all of the $\Tilde{I}_1(Z;S)$ in ascending order, and select the one associated with the representation model that achieves the best downstream prediction performance with the $k$-th percentile ($k$ is a hyperparameter).
(3) Let $\gamma + \upsilon$ represent the difference between the chosen $\Tilde{I}_1(Z;S)$ and $\psi(\epsilon)$.
Suppose none of the tried settings of L-MIFR achieves $\Delta_{\text{DP}}(\tau, \phi) \le \epsilon$, let \proj return \texttt{NSF}.
We introduce the hyperparameter $c$ to prevent overly conservative estimation of $\gamma + \upsilon$ since $\epsilon - c$ serves as a lower bound for $\Delta_\text{DP}$.
Additionally, we use the $\Tilde{I}_1(Z;S)$ value from the $k$-th percentile, rather than the smallest value, to estimate $\gamma + \upsilon$ because $\Tilde{I}_1(Z;S)$ also serves as an upper bound for $I(X;Z|S)$ (Appendix~\ref{apx:migap}).
To maintain the high expressiveness of the representation models, it is essential that $I(X;Z|S)$ remains relatively large. Therefore, we estimate $\gamma + \upsilon$ using one of the smallest $\Tilde{I}1(Z;S)$ values that simultaneously satisfies $\Delta{DP}(\tau, \phi) \le \epsilon$ and avoids excessively reducing $I(X;Z|S)$, which could lead to poor prediction performance.
Overall, we define \begin{align}
 \Tilde{g}'_{\epsilon}(\phi) = \Tilde{I}_1(Z;S) - (\psi(\epsilon) + \gamma + \upsilon),
\end{align}
and we use \proj to ensure that $\Tilde{g}'_{\epsilon}(\phi) \le 0$ with high probability.

%\textcolor{red}{[Phil: This paragraph below is important. There are some wording/grammar issues ,but the message is very good.]}
While this approach is heuristic and may not provide the high-confidence fairness guarantee defined in Def.~\ref{def:highConfFairRep}, \proj can guarantee $\Tilde{I}_1(Z;S) \le \psi(\epsilon) + \gamma + \upsilon$ with high confidence.
We show empirically in Sec.~\ref{sec:exp_practical} that \proj with the practical adjustment in Sec.~\ref{sec:practical}  using this estimation of $\gamma + \upsilon$ tends to satisfy $\Delta_\text{DP}(\tau, \phi) \le \epsilon$.% even though the high confidence guarantee no longer exists. 
   
%\TODO{If there are multiple downstream tasks, this may be problematic. We may find $\gamma + \upsilon$ that bounds $\Delta_\text{DP}(\tau, \phi)$ for one task but not the other.}

% the emprical  we can get from running .  It confirms that $\Tilde{I}_1(Z;S) \ge$  and apply the constraint on \proj to find representation models that  in place of the constraint $I(Z;S)$ and evaluate in experiment. Using the \emph{Adult} dataset (details in Sec.~\ref{sec:exp}) , we observe the non-trivial gap between $\Tilde{I}_1(Z;S)$ that achieves $\Delta_{\text{DP}}(\tau, \phi)$ and $\psi(\Delta_{\text{DP}}(\tau, \phi))$, as shown in Fig.~\ref{fig:gap}.
%  %
%  To construct the constraint 
% $\Tilde{I}_1(Z;S) \le \psi(\epsilon) + \gamma + \upsilon$ where we want $\gamma + \upsilon$ to estimate the gap, we propose using the following mechanism to determine $\gamma + \upsilon$.
%
%Suppose one wants \proj to guarantee $\Delta_\text{DP}(\tau, \phi) \le \epsilon$ with high confidence. 

\section{Datasets}
\label{apx:datasets}
\begin{table}[t]

% \vspace{-1.5mm}
\centering \small
\resizebox{0.9\textwidth}{!}{%
\begin{tabular}{r|cccccc}
\hline
\textbf{Datasets}   & \textbf{Sensitive (n groups)} & \textbf{Downstream Tasks} & \textbf{Data Size} & \textbf{Data Fractions of Each Group} & \textbf{$\Delta_{\text{DP}}$ of True Labels} & \textbf{Feature Dimensions}\\ \hline
Adult  & Gender (\textbf{2})& Income & 41034 & 0.668, 0.332 & 0.260 & 117\\ 
UTK-Face     & Ethnicity (\textbf{5}) & Gender \& Age   & 23700 & 0.425, 0.191, 0.145, 0.168, 0.071 & 0.120 \& 0.319  & $48 \times 48$ \\%PR each group. AGE , 0.600, 0.515, 0.314, 0.462, 0.281. GENDER 0.457, 0.488, 0.541, 0.431, 0.551

%Health  & Age (\textbf{9}) & 2 \\
%German & Gender (\textbf{2}) & 1 & 1000 & 0.69, 0.31 & 0.082 & 58
\hline
\end{tabular}%
}
\caption{\small Summary of dataset statistics. $\Delta_{\text{DP}}$ of True Labels is calculated with $|\Pr(Y = 1 | S = 1) -  \Pr(Y = 1 | S = 0)|$ where $Y$ is the true label.}
\label{tab:dataset}
% \vspace{-1.2mm}
\end{table}

The dataset statistics are listed in Table~\ref{tab:dataset}. The first dataset is the UCI \emph{Adult} dataset,\footnote{\url{ https://archive.ics.uci.edu/ml/datasets/Adult}} which contains information
of over 40,000 adults from the 1994 US Census.
The
downstream task is to predict whether an individual
earns more than \$50K/year with gender as the sensitive attribute.

The second dataset is \emph{UTK-Face},\footnote{\url{https://susanqq.github.io/UTKFace/}} which is a large-scale face dataset with over 20,000 face images with annotations of age, gender, and ethnicity.
We consider ethnicity as the sensitive attribute, and we include two downstream binary prediction tasks: predicting gender and classifying age groups (above or below 30).
We use the pre-processed data from Kaggle.\footnote{\url{https://www.kaggle.com/datasets/nipunarora8/age-gender-and-ethnicity-face-data-csv?resource=download}} 

\section{Baselines}
\label{apx:baselines} 
We include four baselines, among which three are competitive fair representation learning methods. We do not consider baselines that require supervision with a labelled downstream task, such as models proposed by~\citet{Madras2018Learning, Gupta2021Controllable}, etc. We also do not include baselines that restrict the choices of the representation models and the downstream models, such as models proposed by~\citet{Kim2020Fair, Balunovic2022Fair, Jovanovic2023FARE}. We list the baselines we include for experiment with descriptions as follows:
\begin{enumerate}
\item\emph{L-MIFR~\citep{Song2019controllable}} uses Lagrangian Multipliers to encourage a representation model to satisfy constraints $\Tilde{I}_1(Z;S) \le \epsilon_1$ and $\Tilde{I}_2(Z;S) \le \epsilon_2$ where $\Tilde{I}_2(Z;S)$ is an upper bound on $I(Z;S)$ that involves adversarial training(as discussed in Appendix~\ref{apx:upperbounds}), $\epsilon_1$ and $\epsilon_2$ are hyperparameters.%~\citep[Section 2.3]{Song2019controllable} 
%For fair comparison, we fix $\epsilon_1 = \psi(\epsilon) + \gamma + \upsilon$ and tune $\epsilon_2$ only.
%\item\emph{L-MIFR without adversarial training (L-MIFR-no-adv).} This is the same as L-MIFR except it does not consider constraint $\Tilde{I}_2(Z;S) \le \epsilon_2$. This baseline can also be considered as an ablation to \proj with constraint $\Tilde{I}_1(Z;S) \le \psi(\epsilon) + \gamma + \upsilon$, which removes the design of candidate selection and fairness test.
\item\emph{ICVAE~\citep{Moyer2018Invariant}} is a baseline that adds a regularization term $\alpha \Tilde{I}_3(Z;S)$ to the primary loss, where $\alpha \ge 0$ is a hyperparameter, and $\Tilde{I}_3(Z;S)$ is an upper bound on $I(Z;S)$ (as discussed in Appendix~\ref{apx:upperbounds}).
\item\emph{VFAE~\citep{Louizos2016Variational}} is a baseline that adds an  Maximum Mean
Discrepancy (MMD) regularizer, which encourages statistical independence between $S$ and $Z$.
\item Finally, we include the vanilla \emph{VAE}, which is trained solely on the proposed primary objective (Sec.\ref{sec:objective}) without the constraint and does not include extra consideration for fairness.
\end{enumerate}
\section{Hyperparameter Tuning}
\begin{figure}
\centering
\includegraphics[width=0.45\textwidth]{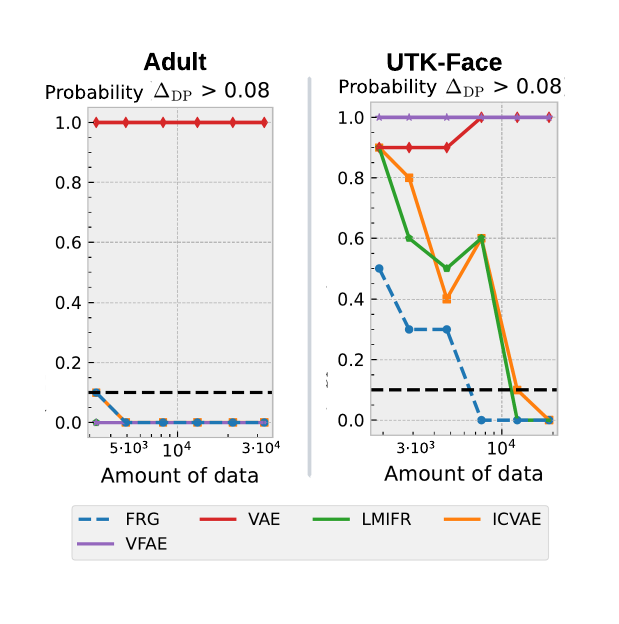}
\vspace{-3.7mm}
\hspace{-3mm}
\caption{\small{
We show the probabilities that $\Delta_{\text{DP}}(\tau,\phi) > 0.08$ evaluated on the \emph{validation} sets (or fairness test set for \proj) for both datasets. On the Adult dataset, all baselines except VAE can maintain the probability that $\Delta_{\text{DP}}(\tau,\phi) > 0.08$ to be less than $\delta = 0.1$. On the UTK-Face, when the data size is relatively large (65\% and 100\% of the trianing data), we find a set of hyperparameters for \proj, LMIFR, ICVAE that keeps $\Pr(\Delta_{\text{DP}}(\tau,\phi) > 0.08) \le \delta$. However, when the data size is small, it is difficult for all methods to keep the probability small on the validation sets.
}}
\label{fig:exp_valid}
\vspace{-3.0mm}
\end{figure}
\label{apx:hyperparam} In our hyperparameter tuning process, we adjust various parameters, including the step sizes of the primary objective, the Lagrange multiplier, and the adversary (for L-MIFR), the weight of the regularizers (e.g. MMD for VFAE), the number of epochs, etc.
The primary objective of hyperparameter tuning is not only to find a set of hyperparameters for the algorithm that minimizes $\Delta_{\text{DP}}$.
Instead, our goal is to find hyperparameters that allow the algorithm to consistently provide a representation model that is $\epsilon$-fair with as high expressiveness as possible.
Thus, one should not tune hyperparameters separately for each of the training datasets we created. When we reuse the same training or validation set for hyperparameter search, we end up evaluating $\Delta_{\text{DP}}$ multiple times on the same training or validation set. As a result, $\Delta_{\text{DP}}$ evaluated on the model trained with the chosen hyperparameters may provide a biased estimation of $\Delta_{\text{DP}}$ on unseen future data.
Consequently, the estimation of the probability $\Delta_{\text{DP}} \le \epsilon$ will also be biased. 
Therefore, we create additional datasets for hyperparameter tuning and adopt the same hyperparamters on different training datasets of the same size.
%
% We create 10 additional datasets for each data sizes by sampling with replacement from the non-test data, which we call the \emph{tuning datasets}.
%
% We hold out 20\% of each tuning dataset as the \emph{validation set}.
%

For baselines, we create validation sets by sampling 20\% of the training data, while for \proj, we evaluate the models using the fairness test datasets (i.e., $D_f$ in Sec.~\ref{sec:ft}). We tune each algorithm with grid search according to the metrics evaluated on the validation sets (for baselines) or on the fairness test sets (for \proj). For the UTK-Face dataset, as there are multiple downstream tasks, we only assume the gender labels are available for hyperparameter tuning.
We first consider hyperparameters that yield high probabilities (at least $1-\delta$) of satisfying $\Delta_{\text{DP}} \le \epsilon$ .
If the probabilities of satisfying $\Delta_{\text{DP}} \le \epsilon$ for all sets of hyperparameters are lower than $1-\delta$, we select the sets that provide the highest probabilities while maintaining the lowest average $\Delta_{\text{DP}}$.
%
%Among those sets, for \proj, we select the ones with the lowest estimated probabilities of encountering \texttt{NSF}.
%
If there are ties, we prioritize the hyperparameters that achieve the highest average AUC.
Note that we set the minimum allowed step size for the primary objective to $10^{-6}$. This choice is motivated by the fact that an algorithm with an excessively small step size may have minimal impact on optimizing the primary objective and could potentially produce random representations that lack utility for downstream predictions, despite being highly likely to be fair. We also note that we use the same number of dimensions for representation $Z$ and the same hidden size for the downstream MLP for fair comparison. We show the resulting $\Pr(\Delta_{\text{DP}}(\tau,\phi) > 0.08)$ evaluated on the validation sets (or fairness test set for \proj) for both the Adult and the UTK-Face datasets in Fig.~\ref{fig:exp_valid}.

\section{Ablation Study}\label{apx:ablation}
\begin{figure}
\centering
\includegraphics[width=0.8\textwidth]{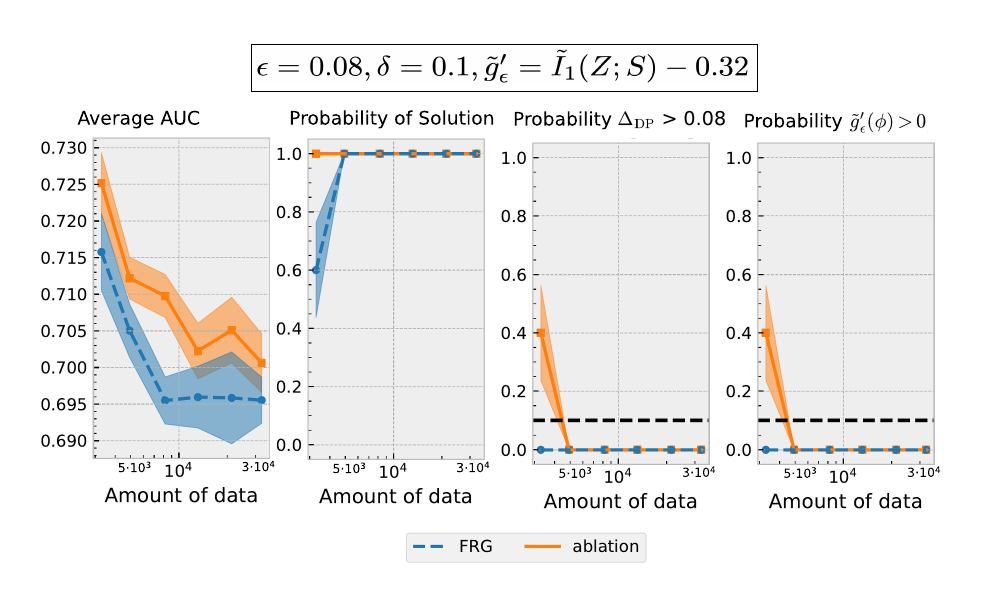}
\vspace{-3.7mm}
\hspace{-3mm}
\caption{\small{
Ablation study on \proj with the practical adjustments (Sec.~\ref{sec:practical}) using the adult dataset. We include the fourth plot which shows the probabilities the constraint $\Tilde{g}'_\epsilon(\phi) \le 0$  is violated. See Appendix~\ref{apx:ablation} for discussion.
}}
\label{fig:exp_ablation}
\vspace{-3.0mm}
\end{figure}

\begin{figure}
\centering
\includegraphics[width=0.8\textwidth]{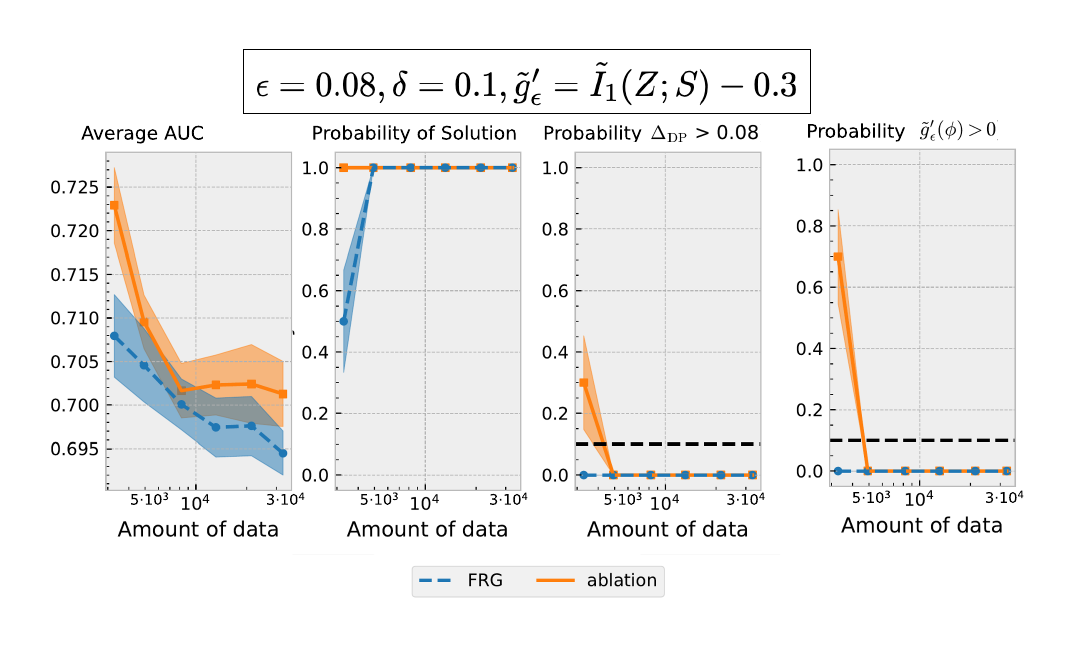}
\vspace{-3.7mm}
\hspace{-3mm}
\caption{\small{
Ablation study on \proj with the practical adjustments (Sec.~\ref{sec:practical}) using the adult dataset. Different from Fig.~\ref{fig:exp_ablation}, it uses a slightly stricter constraint $\Tilde{g}'_\epsilon = \Tilde{I}_1(Z;S) - 0.3$.  See Appendix~\ref{apx:ablation} for discussion.
}}
\label{fig:exp_ablation_strict}
\vspace{-3.0mm}
\end{figure}

We conduct an ablation study on  \proj with the practical adjustments (Sec.~\ref{sec:practical}) with two changes: (1) removing the fairness test component, (2) modifying the candidate selection process. Instead of using $\hat{U}_\epsilon(\phi,D_c) \le 0$ as an optimization constraint, where $\hat{U}_\epsilon(\phi,D_c)$ is a predicted high-confidence upper bound on $\Tilde{g}'_\epsilon(\phi)$ (Sec.~\ref{sec:cs}), we only evaluate the expectation of $\Tilde{g}'_\epsilon(\phi)$ using candidate selection data $D_c$.
%and the prediction of the high-confidence upper bound on $\Tilde{g}'_\epsilon(\phi)$ (i.e. the construction of $\hat{U}_\epsilon(\phi,D_c)$) in candidate selection (Sec.~\ref{sec:cs}). We use the mean of  
%
Formally, the ablation solves the following constrained optimization problem using the KKT conditions: 
\begin{align}
    \max_{\theta, \phi}~~ & 
 \mathbb{E}_{q_\phi(Z|X,S)}\Big [\log p_\theta(X|Z,S)\Big ] - \mathbb{KL}\Big(q_\phi(Z|X,S) \| p(Z)\Big)\\
    \text{s.t. }~~ & \ \mathbb{E}\Big [\Tilde{g}'_\epsilon(\phi)\Big ] \le 0.\label{eq:ablation_constraint}
\end{align}
We conduct the experiment on the Adult dataset, and we show the results in Fig.~\ref{fig:exp_ablation}.% and Fig.~\ref{fig:exp_ablation_strict}. Note that Fig.~\ref{fig:exp_ablation} uses the same $\Tilde{g}'_\epsilon(\phi)$ as the experiment in ~Sec.\ref{sec:exp_practical} while Fig.~\ref{fig:exp_ablation_strict} uses a slightly stricter constraint $\Tilde{g}'_\epsilon = \Tilde{I}_1(Z;S) - 0.3$.

The results indicate that, although the ablated \proj achieves a slightly better AUC, it can produce representation models that do not meet the $\epsilon$-fairness requirement due to a lack of consistent constraints on $\Tilde{g}'_\epsilon(\phi) \le 0$ with high probabilities.
This highlights the necessity of a fairness test that assesses the high-confidence upper bounding of $\Tilde{g}'_\epsilon(\phi)$ to ensure $\epsilon$-fairness in representation models.

We also observe that the original method has the probabilities of returning \texttt{NSF} (the second plot) that are equivalent to the ablated method's probabilities of violating the constraint $\Tilde{g}'_\epsilon(\phi) \le 0$ (the fourth plot).
Including the fairness test for the ablated method would result in the same probability of returning \texttt{NSF} as the original method. This demonstrates that using the constraint $\mathbb{E}\Big [\Tilde{g}'_\epsilon(\phi)\Big ] \le 0$ and the constraint $\hat{U}_\epsilon(\phi,D_c)$ for optimization in candidate selection has a similar effect in this experiment.

We use $\hat{U}_\epsilon(\phi,D_c)$ because previous work~\citep{Philip2019Preventing, Metevier2022Offline,Hoag2023Seldonian} has demonstrated its effectiveness in preventing overfitting to training data when aiming to satisfy constraints with high probability on ground truth data.
To assess whether using $\hat{U}\epsilon(\phi,D_c)$ provides an advantage, we evaluate \proj using a more stringent constraint, $\Tilde{g}'\epsilon(\phi) \coloneqq \Tilde{I}_1(Z;S) - 0.3$, as shown in the results in Fig.~\ref{fig:exp_ablation_strict}.
As expected, with the smallest data size, because of a stricter constraint, \proj is more likely to return \texttt{NSF} (the second plot) while the ablated \proj achieves a lower probability of $\Delta_{\text{DP}}(\tau,\phi) > 0.08$ (the third plot).
%
% \textcolor{red}{AH: I'm not sure what you're trying to say in this next sentence. Maybe just needs a rewording.} 
However, we observe on the fourth plot that the ablated \proj violates $\Tilde{g}'_\epsilon(\phi) \le 0$ with a large probability, even larger than the probability that the original \proj returns \texttt{NSF}.
This observation may be explained by the hypothesis that using the constraint $\mathbb{E}\Big [\Tilde{g}'_\epsilon(\phi)\Big ] \le 0$ can lead to overfitting on the training data and does not provide a high probability that the constraint $\Tilde{g}'_\epsilon(\phi) \le 0$ will be satisfied on future unseen data.
For future work, we will investigate whether there can be better alternative constraints than $\hat{U}_\epsilon(\phi,D_c) \le 0$ in candidate selection to find candidate solutions that are likely to pass the fairness test while achieving better expressiveness.

\end{document}